# Observing Health Outcomes Using Remote Sensing Imagery and Geo-Context Guided Visual Transformer

Yu Li, *Member, IEEE*, Guilherme N. DeSouza, *Senior Member, IEEE*, Praveen Rao, *Senior Member, IEEE*, Chi-Ren Shyu, *Senior Member, IEEE*

***Abstract*—Visual transformers have driven major progress in remote sensing image analysis, particularly in object detection and segmentation. Recent vision-language and multimodal models further extend these capabilities by incorporating auxiliary information, including captions, question and answer pairs, and metadata, which broadens applications beyond conventional computer vision tasks. However, these models are typically optimized for semantic alignment between visual and textual content rather than geospatial understanding, and therefore are not well suited for representing or reasoning with structured geospatial layers. In this study, we propose Geo-Context Guided Visual Transformer that enhances remote sensing imagery processing with auxiliary geospatial guidance. The proposed approach introduces a geospatial embedding mechanism that converts heterogeneous geospatial variables into patches-aligned representations and an asymmetric geo-context guided attention module that uses structured geospatial context to modulate visual attention while preserving the input-centered representation stream. The module also assigns geospatial roles to attention heads, supporting structured interpretation of image-geospatial interactions. Experimental results show that the proposed framework outperforms pretrained remote-sensing vision-language models and graph-based spatial fusion baselines in disease prevalence prediction. Ablation and visualization analyses further indicate its value for health-related remote sensing tasks where comprehensive geospatial data may be limited, while providing interpretable spatial cues for subsequent public-health analysis.**

Yu Li is with the Department of Electrical Engineering and Computer Science, University of Missouri, Columbia, MO 65201 USA (e-mail: ylrbc@missouri.edu).

Guilherme N. DeSouza is with the Department of Electrical Engineering and Computer Science, University of Missouri, Columbia, MO 65201 USA (e-mail: desouzag@missouri.edu).

Praveen Rao is with the Department of Electrical Engineering and Computer Science, University of Missouri, Columbia, MO 65201 USA (e-mail: praveen.rao@missouri.edu).

Chi-Ren Shyu, corresponding author, is with the Department of Electrical Engineering and Computer Science, and the Institute for Data Science and Informatics, University of Missouri, Columbia, MO 65201 USA (e-mail: shyuc@missouri.edu).

## I. INTRODUCTION

RECENT advances in high-resolution remote sensing imagery and deep neural networks have substantially improved tasks such as scene classification [1], [2], [3], object detection [4], [5], and land cover segmentation [6], [7], [8]. However, these tasks are typically formulated as image-only problems, where the main predictive signal is assumed to lie largely within the image footprint. This assumption becomes limiting when outcomes depend on contextual factors beyond the visible scene, particularly in public-health applications where interpretability is also important.

Regional health outcomes are shaped by socioeconomic conditions [9], [10], [11], environmental exposures [12], [13], [14], [15], and access to healthcare [16], [17], [18]. Although some of these determinants may be partially reflected in the built and natural environment visible in remote sensing imagery [19], [20], [21], [22], [23], [24], many remain outside the image or are not directly observable. As a result, image-only models are limited in their ability to capture the broader contextual drivers of health outcomes. This limitation is compounded by increasing model complexity, which can make it difficult to determine whether predictions are driven by meaningful environmental patterns or by spurious visual correlations.

Existing multimodal approaches have extended remote sensing models beyond image-only inputs, particularly through vision-language alignment frameworks [25], [26], [27], [28]. These methods have shown strong capabilities in linking imagery with language-based supervision closely tied to visible content. However, in applications such as health outcome prediction, auxiliary information often consists of high-dimensional structured geospatial variables, including socioeconomic, demographic, environmental, and infrastructural contextual layers. These variables can capture

The supplemental document provides model configuration comparison, detailed results for Section V-A and examples for generated text summaries.

This article has been published as:

Y. Li, G. N. DeSouza, P. Rao, and C.-R. Shyu, "Observing Health Outcomes Using Remote Sensing Imagery and Geo-Context Guided Visual Transformer," *IEEE Transactions on Geoscience and Remote Sensing*, 2026, doi: 10.1109/TGRS.2026.3725585.

effects that are indirect, spatially heterogeneous, or not visually explicit. This setting differs from conventional image-text alignment and motivates approaches that can directly incorporate structured geospatial context while preserving interpretable links between contextual variables and visual evidence. Beyond this modeling challenge, comprehensive collections of such variables can be costly to assemble consistently and may be incomplete where local data infrastructure or standardized records are limited. A detailed review of related transformer, multimodal, and health-prediction studies is provided in Section II.

To address this gap, we propose a Geo-Context Guided Visual Transformer that integrates geospatial contextual layers into visual transformer architectures through guided attention and geospatial embedding. The model uses auxiliary geospatial information to modulate attention during feature extraction, enabling task-relevant focus while preserving a clear separation between input and guidance. The main contributions of this work are as follows:

1) A geospatially conditioned, asymmetric attention module that uses structured geospatial variables to modulate visual attention weights and head contributions while preserving the primary input stream.
2) A geospatial embedding mechanism that transforms heterogeneous geospatial variables into spatially aligned embedding patches for multimodal representation learning.
3) An interpretable multimodal design that separates input from guidance, enabling attention-based analysis of how contextual variables influence prediction.
4) A practical demonstration that accurate health-outcome prediction is feasible using remote sensing imagery and a limited set of geospatial variables, reducing reliance on extensive census data while revealing interpretable spatial associations relevant to public health and urban management.

## II. Background

### A. Vision Transformers with Remote Sensing Images

Since the introduction of the Vision Transformer (ViT) [29] for image classification, transformer-based architectures have become widely adopted across computer vision, including in remote sensing. Compared with natural images, remote sensing imagery presents unique challenges, such as top-down perspective, large-scale spatial structures, and distinct object types and distributions. These differences motivate architectural adaptations tailored to this domain.

*1) Scene Classification:* Early studies demonstrated that ViT pretrained on natural images can perform competitively with CNNs for scene classification tasks in remote sensing [30]. Subsequent work explored training strategies to improve adaptation, including generative self-supervised learning [31], [32], multi-task learning [33], and contrastive learning frameworks such as ViT-CL [34]. Hybrid architectures combining CNNs and transformers have also been proposed, including dual-stream encoders (CTNet [35], L2RCF [36]), ResNet–Transformer fusion models (Resformer [37], TRS [38]), and multi-scale patch designs (LG-ViT [39]). Other efforts such as TSTNet [40] incorporated edge-enhanced branches to address noise and data scarcity.

Beyond training approaches, several methods introduced architectural modifications specific to remote sensing. HHTL [41] adjusted patch generation to improve color consistency; DSS-TRM [42] and SpectralFormer [43] tailored transformer designs to hyperspectral data; MITformer [44], EMTCAL [45], and Yang et al. [46] integrated convolutional operations or enhanced attention mechanisms to encode local structure and multi-scale features. Despite these innovations, relatively few studies have focused on modifying attention mechanisms to incorporate external or auxiliary modalities.

*2) Object detection:* Transformer-based object detection in remote sensing often employs hybrid CNN-transformer designs to capture both local detail and global spatial context. Methods such as TRD [47], GOCPK [48], and CI_DETR [49] use CNN-derived feature pyramids as transformer inputs. Architectures including LPSW [50], EIA-PVT [51], LAST-Net [52], and the model by Lu et al. [53] incorporate convolutional blocks within or alongside transformer layers to strengthen local feature extraction, which is critical for detecting small, densely clustered objects like vehicles.

Other work focused on improved learning strategies or attention designs. Zhang et al. [54] applied adversarial and self-supervised pretraining; Zheng et al. [55] introduced the Feature Pyramid Transformer (FPT) to propagate attention across feature scales; and Yuan and Wei [56] leveraged cross-attention between RGB and infrared channels to improve multi-spectrum detection accuracy. Overall, object detection methods highlight the need for architectures that combine transformers' global reasoning with mechanisms that handle domain-specific spatial structure.

### B. Multimodal Models with Remote Sensing Images

*1) Modality-Separate Encoder Models:* Large vision-language models (VLMs) such as CLIP [25] and ERNIE-ViL [57] have motivated multimodal learning in remote sensing, particularly for image-text retrieval. Early systems paired transformer-based image and text encoders (e.g., ViT and Transformer in [58]; SIRS [59]) and performed alignment by comparing features extracted from each modality independently. MSCMAT [60] extended this paradigm by introducing a multiscale alignment strategy for refined image-text feature matching. CLIP-based adaptations later became dominant: PERT-RaMa [61] applied parameter-efficient tuning, CUP [62] introduced token generation while freezing CLIP weights, and methods such as CLGSA [63] and PR-CLIP [64] added attention refinements to strengthen image-text alignment within the dual-encoder design.

Recent efforts scaled these approaches using large remote sensing datasets and related alignment objectives. RemoteCLIP [26] leveraged diverse detection, segmentation, and text-image datasets; GeoRSCLIP [27] and SkyScript [28] compiled multimillion image-text pairs for CLIP-style training; CLIP-MoA [65] modified CLIP with gating for multi-task learning;

EarthGPT [66] combined CNN and ViT image encoders with a Llama-based text model; and SatCLIP [67] learned general-purpose location embeddings by aligning satellite imagery with geographic coordinates. Most of these models maintain dual-encoder CLIP-style pipelines, benefiting from natural language supervision but still bound by token limits and language-model constraints.

These constraints pose challenges for incorporating high-dimensional geospatial layers. CLIP's typical token budgets (77 for CLIP; 512–1024 for extended variants [68], [69]) cannot accommodate hundreds of geospatial variables unless expanded into long natural-language descriptions that exceed encoder capacities. Moreover, text inputs in VLMs usually describe visible scene content, whereas many geospatial variables capture latent socioeconomic, infrastructural, or environmental attributes that are not directly observable in imagery. As a result, standard multimodal VLMs struggle to model the weak and indirect correspondence between remote sensing imagery and structured geospatial data, motivating alternative cross-modal integration mechanisms.

*2) Guided Attention for Cross-Modal Interaction:* Guided attention generally refers to attention mechanisms that condition visual feature weighting on auxiliary inputs rather than relying solely on image-derived self-attention. Prior to the prevalence of transformers, early work such as GAIN [70] demonstrated that externally supervised attention can improve visual localization by steering attention toward semantically relevant regions. Text-conditioned variants later enabled language-guided visual attention for tasks such as image captioning [71], and language-specified attention supervision has been used to encourage image classifiers to attend to task-relevant visual evidence under biased or noisy conditions [72]. With the development of transformer-based multimodal learning, cross-modal attention has been widely studied in vision-language models. ViLBERT [73] and LXMERT [74] process visual and linguistic tokens in separate streams and exchange information through co-attentional or cross-modality layers, whereas UNITER [75] learns joint image-text embeddings with a shared transformer over both modalities. Similar ideas have also been explored in image-plus-metadata settings outside remote sensing. Pacheco and Krohling [76] introduced an attention-based mechanism to combine skin lesion images with patient metadata, and MCAT [77] used genomic-guided co-attention to model interactions between histology patches and gene embeddings for survival prediction. Recent multimodal systems further extend guided attention by injecting textual, relational, or structural cues into attention modules for tasks such as financial forecasting [78], vision-language robustness [79], and chart question answering [80].

In remote sensing, guided and cross-modal attention have also appeared in several forms. One line of work uses task-specific guidance, including segmentation masks for visual question answering [81], category-aware cues for semantic segmentation [82], and gradient-structure guidance for generative modeling [83]. Another line focuses on multimodal sensor fusion or representation learning [84]. MFT [85] was proposed for HSI-LiDAR classification using cross-patch interaction between complementary modalities. Liu et al. [86] introduced a transformer-based multimodal fusion network that incorporates multispectral imagery and DSM height information for semantic segmentation. CROMA [87] learns optical-SAR representations from spatially aligned multimodal inputs using contrastive and masked reconstruction objectives, while SatMIP [88] aligns satellite images with acquisition metadata represented as captions. Thus, existing remote-sensing methods mainly treat auxiliary information as co-registered sensor measurements, acquisition metadata, masks, labels, or structural priors, and are less concerned with guidance-conditioned attention over heterogeneous socioeconomic, demographic, and environmental variables.

Graph neural networks provide another relevant approach for modeling structured spatial context, particularly when image patches, geographic units, or spatial regions can be represented as nodes connected by spatial or semantic relationships. Graph attention variants, such as GATv2 [89], enable adaptive aggregation from neighboring nodes, while graph transformer architectures, such as GraphGPS [90], extend this idea by combining local message passing with mechanisms for broader-range dependency modeling. These architectures are suited for capturing spatial relationships among geographic entities. However, when applied directly to multimodal geospatial prediction, visual evidence and auxiliary contextual variables are often represented within a shared node representation. Such fusion may improve predictive modeling of spatial dependencies, but it can also blur the distinction between image-derived evidence and geospatial context. As a result, graph attention weights may help characterize relationships among regions, but they might not explain how specific geospatial categories interact with visual features.

The guidance used in our setting consists of high-dimensional, continuous, and heterogeneous geospatial layers whose relationships with image evidence are indirect, spatially variable, and task-dependent. Instead of treating imagery and guidance as symmetric modalities for generic alignment or sensor fusion, the proposed method preserves the remote sensing image as the primary representational stream and uses geospatial information to modulate attention weights and head contributions. Additionally, by assigning geospatial categories to distinct attention heads, the model further provides category-specific attention maps that help relate visual regions to interpretable socioeconomic, demographic, and environmental factors.

### *C. Health Outcome Prediction with Remote Sensing Images*

Compared with traditional remote sensing tasks, predicting health outcomes from imagery remains limited. Health-oriented geospatial research has linked spatial context to population-health analysis, but building comprehensive contextual resources requires substantial integration of heterogeneous social, environmental, and infrastructural information across sources and geographic units [91]. In the remote-sensing literature, most prior work extracts intermediate features and factors rather than training end-to-end predictors. Remote sensing images have been employed to estimate health-relevant factors such as fire plumes [92], water tanks [93], livestock distribution [94], heavy metal concentrations [95], [96], measures of environmental deprivation [97], [98], and broader socioeconomic indicators [99]. Extracted features have then been linked to health outcomes through downstream models, such as obesity prediction using VGG features [100], cancer prevalence prediction using ResNet features [101], and COVID-19 risk factor prediction with AlexNet and XGBoost [102]. Other studies, such as [103], employed clustering of extracted image features to analyze associations with urban health indicators.

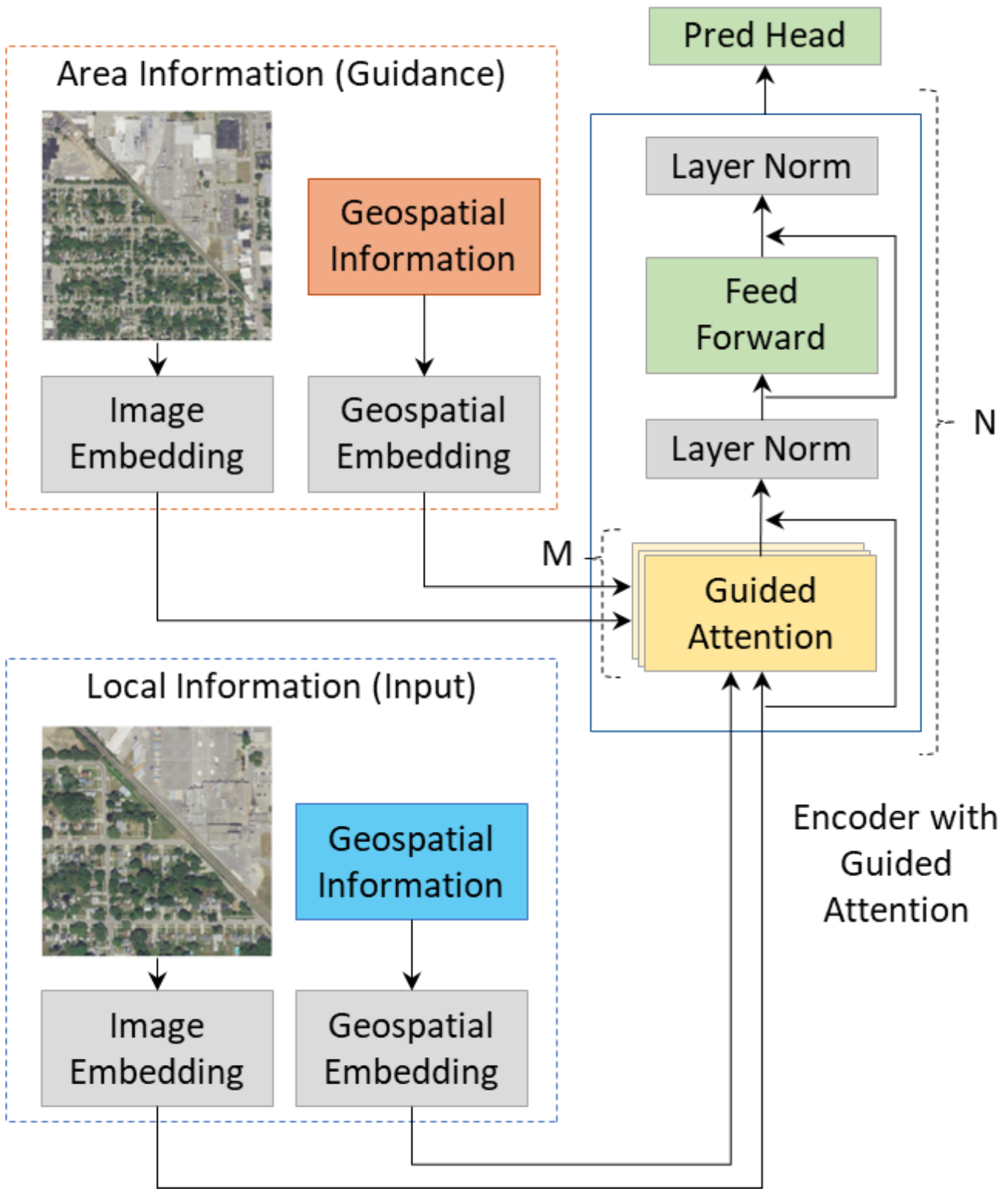

**Fig. 1.** Overall architecture of a variant of the proposed Geo-Context Guided Visual Transformer. Geospatial information is processed into Geospatial Embeddings and integrated with the image embedding in the proposed Guided Attention. Guidance information regulates the input processing through the generation of head and attention weights. The encoded features are subsequently passed to a prediction head to produce the final health outcome prediction.

Only a small number of works perform direct end-to-end prediction of health outcomes. Levy et al. [104] used transfer learning with a pretrained ResNet to predict mortality rates from imagery sampled around schools, illustrating the feasibility of direct prediction but leaving the design of multimodal health-focused models largely unexplored.

Despite the rapid progress of transformer-based architectures in remote sensing reviewed in Section II, existing models primarily focus on visual information extracted from imagery alone, often overlooking the rich geospatial context that governs many real-world phenomena. While recent variants of visual transformers have incorporated architectural enhancements, these efforts remain constrained to visual features and lack explicit mechanisms for integrating external geospatial data. Similarly, multimodal approaches inspired by vision–language models like CLIP have shown promise in aligning heterogeneous modalities, but their reliance on text encoders and fixed token capacities limits their applicability to high-dimensional geospatial layers. Moreover, such models are typically optimized for tasks with direct visual-semantic correspondence, whereas geospatial variables often represent latent environmental, socioeconomic, or infrastructural conditions that are not visually explicit. This gap is particularly critical in emerging applications such as health outcome prediction, where spatially distributed factors influence observable imagery in complex and indirect ways.

Therefore, there is a pressing need for a framework that goes beyond generic modality alignment or co-registered sensor fusion by using structured geospatial variables as contextual guidance for visual attention, while preserving interpretable links between geospatial categories, image regions, and downstream health outcomes.

## III. Methods

In this section, we present an interpretable framework with the Geo-Context Guided Visual Transformer (GCGVT) to effectively incorporate multimodal contextual information for health outcome prediction using remote sensing imagery.

### *A. Overview*

We propose two variants of GCGVT, GCGVT-A and GCGVT-G, each motivated by a distinct design hypothesis presented in Section III-D. Fig. 1 presents the architecture of the first variant, which incorporates area-level contextual guidance in conjunction with local data. Geospatial information is first transformed into spatially aligned geospatial embeddings and then used by the Guided Attention (GA) module to condition the processing of image embeddings. As shown in Fig. 1, the model combines these multimodal features through GA to produce a unified representation, which is subsequently passed to a prediction head for health outcome estimation.

At the core of both architectures is the GA (Fig. 2), a geospatially conditioned guided-attention module. Unlike generic co-attention or cross-attention modules that jointly update peer modalities, GA is designed as an asymmetric conditioning mechanism: the primary input, such as a local remote sensing image, remains the value-carrying stream, while the auxiliary geospatial context modulates attention weights and attention-head contributions. In addition, the architecture enforces distinct attention heads to specialize in different aspects of the guidance information, thereby improving the explainability of the resulting predictions.

### *B. Guided Attention*

The proposed GA module extends multi-head attention by introducing geospatial guidance into both patch-level attention computation and head-level modulation. The purpose of GA is not to learn a generic joint representation between two peer modalities. Instead, it preserves the primary input as the main value-carrying stream and uses auxiliary geospatial information to guide how that input is attended. This asymmetric formulation is motivated by remote-sensing-based health prediction, in which the image provides spatially localized visual evidence, whereas geospatial variables provide contextual factors that may be only weakly or indirectly expressed in the image.

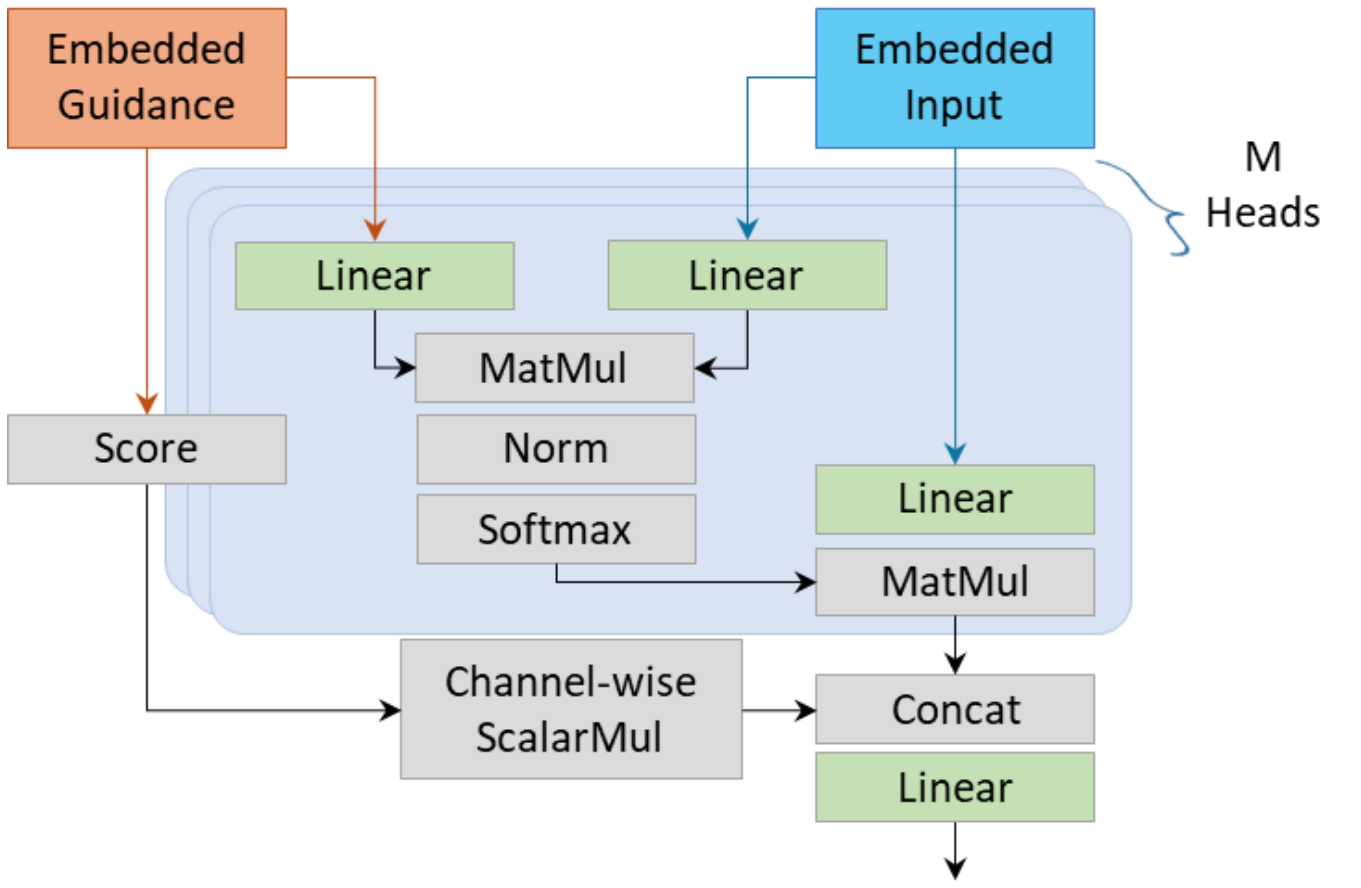

**Fig. 2.** Architecture of the Guided Attention module, showing how auxiliary guidance information is integrated with input embeddings. The embedded guidance (orange) is linearly projected and used to generate a modulation score through a sigmoid function, while the embedded input (blue) undergoes standard attention operations. The guidance score adjusts the attention output via channel-wise scaling before concatenation and linear transformation, producing the fused multimodal representation.

Compared with standard self-attention, in which the query, key, and value are derived from the same input sequence, GA derives the key and value from the primary input and the query from the guidance input. Compared with generic cross-attention, GA further introduces a guidance-derived head-weight vector that modulates the contribution of individual attention heads. This design allows each head to represent a specific aspect of the geospatial guidance, rather than allowing all heads to mix guidance categories without semantic structure. As illustrated in Fig. 2, the proposed module uses guidance information as a reference for generating attention weights and head weights, enhancing relevant regions or features in the primary input while keeping the guidance pathway explicitly distinguishable from the input pathway.

Let $X_I$ denote the position-aware embedded input stream and $X_G$ denote the position-aware embedded guidance stream. $X_I$ and $X_G$ are modality-agnostic and may each contain image embeddings, geospatial embeddings, or their combination, depending on the model. An important property of GA is that the value-carrying stream is derived from $X_I$, while $X_G$ affects the attention weights and the contribution of individual attention heads. For an attention layer with $H$ heads, the $i$-th head computes its key and value from the input stream and its query from the guidance stream:

$$K_i = \psi_K^i(X_I), \quad V_i = \psi_V^i(X_I), \quad Q_i = \psi_Q^i(X_G), \tag{1}$$

where $\psi_K^i(\cdot)$, $\psi_V^i(\cdot)$, and $\psi_Q^i(\cdot)$ are head-specific learnable projections. The attention map and head output are then computed as:

$$A_i = \mathrm{softmax}\left(\frac{Q_i K_i^{\top}}{\sqrt{d_h}}\right), \qquad Y_i = h_i A_i V_i, \tag{2}$$

where $d_h$ is the head dimension and $h_i$ is a scalar head-modulation weight. The outputs of all heads are concatenated and projected to produce the GA output:

$$Y = \mathrm{Concat}(Y_1, \ldots, Y_H)\, W_O. \tag{3}$$

This formulation preserves $X_I$ as the value-carrying stream because $V_i$ is computed only from the input stream. The guidance stream $X_G$ does not replace the input representation. Instead, it determines how the input stream is attended through $Q_i$ and how much each head contributes through $h_i$. Therefore, GA is an asymmetric attention mechanism in which the input stream provides the represented content, while the guidance stream controls the attention pattern and head-level modulation.

In an unconstrained cross-attention layer, each attention head could learn a query projection from the complete guidance vector, i.e., $Q_i = W_{Q,i} X_G$. Although such a formulation is more flexible, it makes category-level interpretation non-identifiable. The attention map of head $i$ may depend on an arbitrary mixture of geospatial categories, and the same query representation can be obtained through different reparameterizations of the guidance embedding and projection matrix. Consequently, a head's attention map cannot be uniquely attributed to a specific geospatial factor. This issue is consistent with the broader observation that disentangled representations require explicit inductive biases [105], and with prior cautions that unconstrained attention weights should not automatically be treated as faithful explanations [106].

To support interpretable head specialization, GA allows the guidance stream to be organized into semantically bounded components. Let the guidance roles be indexed by $r \in \{1, \ldots, R\}$. In the proposed geo-context setting, these roles correspond to broad geospatial categories such as age and sex, education, housing, income, and race, but the formulation itself is not restricted to geospatial data. Let $\rho(i)$ denote the semantic role assigned to head $i$. The guidance projection for head $i$ can then be written as:

$$Q_i = X_G^0 W_{Q,0}^i + X_G^{\rho(i)} W_{Q,\rho}^i, \tag{4}$$

where $X_G^0$ is a head-common guidance component and $X_G^{\rho(i)}$ is the role-specific guidance component assigned to head $i$. When no common guidance component is used, the first term is omitted. The guidance can include imagery from a larger area, which is represented by $X_G^0$, or it can be primarily geospatial, so $X_G^{\rho(i)}$ corresponds to the geospatial category assigned to head $i$ while $X_G^0$ is omitted.

Similarly, the input stream may contain a shared input component and optional role-specific components:

$$K_i = X_I^0 W_{K,0}^i + X_I^{\rho(i)} W_{K,\rho}^i, \tag{5}$$

$$V_i = X_I^0 W_{V,0}^i + X_I^{\rho(i)} W_{V,\rho}^i. \tag{6}$$

Here, $X_I^0$ is the head-common input component, such as the local image embedding, and $X_I^{\rho(i)}$ is an optional role-specific input component, such as a local geospatial embedding from the same semantic category. If the input stream does not include role-specific components, the second term in each expression is omitted.

The head-modulation weight $h_i$ is computed from the guidance role assigned to head $i$:

$$h_i = \mathrm{score}_i\left(s_G^{\rho(i)}\right), \tag{7}$$

where $s_G^{\rho(i)}$ is a role-specific summary vector derived from the guidance information, and score is a customizable function that can be adapted to different applications, allowing predefined formulas or desired structures. In this work, it is used to enforce a semantic structure over attention heads, so that different geospatial categories can contribute to distinct heads. This one-to-one category-to-head mapping is introduced primarily as an interpretability-oriented constraint rather than as an assumption of performance optimality. By anchoring each head to a predefined geospatial role, the head weights provide bounded modulation scores for the specific guidance categories, while the corresponding attention maps indicate which input regions are emphasized under each role-anchored guidance signal.

In our implementation, the score function is modeled as a lightweight category-specific gating network rather than as an unconstrained fusion module. The purpose of this network is to estimate the relevance of each geospatial category for modulating its corresponding attention head. Because the relationship between geospatial variables and health outcomes is nonlinear, spatially heterogeneous, and not governed by a known analytic formula, a fixed hand-designed score would impose an overly restrictive assumption. Conversely, using a deeper or shared network over all geospatial categories would increase model flexibility but would weaken the direct interpretability of the head weights. We therefore use a two-layer fully connected network as a minimal nonlinear gating function. This design follows the same principle as feature recalibration and conditional modulation mechanisms, where compact learned networks generate bounded scaling factors to regulate intermediate representations [107], [108].

Specifically, the head weight is computed as:

$$h_i = \sigma\left(w_2^{i\top} \phi\left(W_1^i s_G^{\rho(i)}\right)\right), \tag{8}$$

where $\phi(\cdot)$ is a nonlinear activation function and $\sigma(\cdot)$ constrains the modulation weight to $(0, 1)$. $s_G$ represents area-weighted geospatial attributes for the region covered by guidance. The first layer allows nonlinear interactions among variables within the assigned geospatial role, while the second layer compresses this information into a single scalar gate for the corresponding attention head. Importantly, the gating network is applied independently to each geospatial category. Thus, the two-layer network is used to learn category-specific head modulation, while the one-to-one category-to-head structure preserves the interpretability of the attention maps.

The semantic constraint in GA is imposed through the role-specific branches in (4) and (7). For a role $r' \neq \rho(i)$, the role-specific guidance pathway of head $i$ has no direct dependence

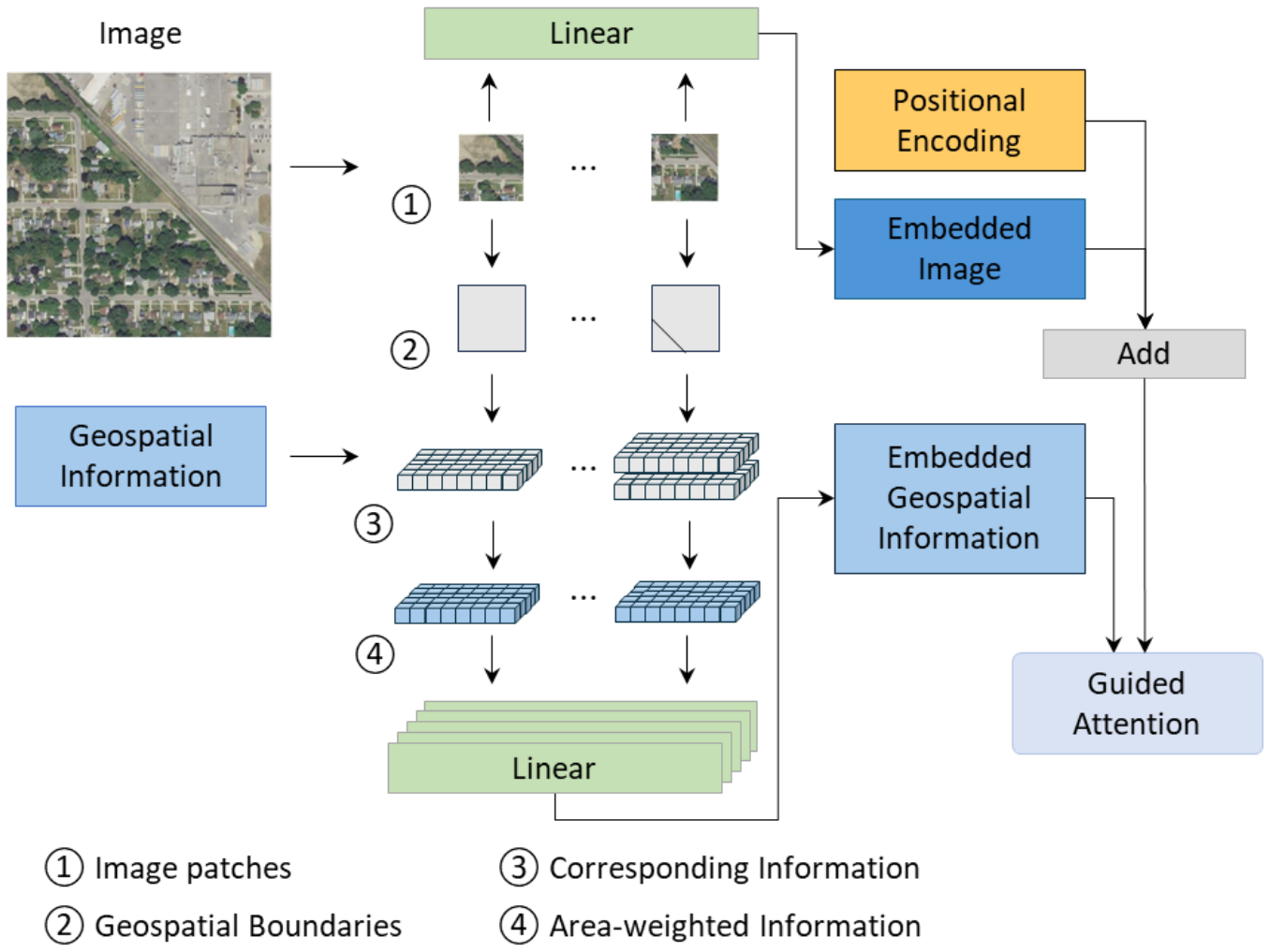

**Fig. 3.** Generation and integration of Geospatial Embeddings. Geospatial information is area-weighted based on image patch boundaries and processed through category-specific linear projections to produce embedded geospatial variables, which are then combined with image embeddings and positional encodings.

on $X_G^{r'}$ or $s_G^{r'}$:

$$\frac{\partial Q_i}{\partial X_G^{r'}} = 0, \qquad \frac{\partial h_i}{\partial s_G^{r'}} = 0, \qquad r' \neq \rho(i), \tag{9}$$

excluding any optional head-common guidance component $X_G^0$. This constraint does not mean that the attention map of head $i$ is determined by a single geospatial category alone. The attention map also depends on the input stream through $K_i$ and $V_i$. Rather, the constraint means that the geospatial guidance and head-modulation pathways are semantically anchored. Each head receives category-preserving geospatial guidance from its assigned role without mixing different geospatial categories before the attention computations.

In this sense, the one-to-one mapping is not claimed to be globally optimal over all possible fusion architectures. It is an interpretation-constrained design choice that trades some cross-category flexibility for a clearer explanation pathway from a geospatial category, to a head-modulation gate, to the corresponding category-guided attention map. The within-category projections and attention weights remain trainable, so the model can still learn input-guidance interactions from data while maintaining category-level traceability.

This design also reflects observations from prior analyses of multi-head attention that different heads can contribute unequally and may specialize in different functions [109], [110]. In GA, this head specialization is not left entirely implicit. Instead, it is guided by geospatial category structure, and the learned head weights quantify the relative contribution of each category-specific guidance signal.

### C. Geospatial Embedding

A key component of the proposed method is the integration of geospatial information through the Geospatial Embedding (Fig. 3). By incorporating an additional modality, the network gains access to contextual information that may not be present in the image itself. This geospatial information also plays a critical role in GA by helping the model differentiate between input data and guidance signals.

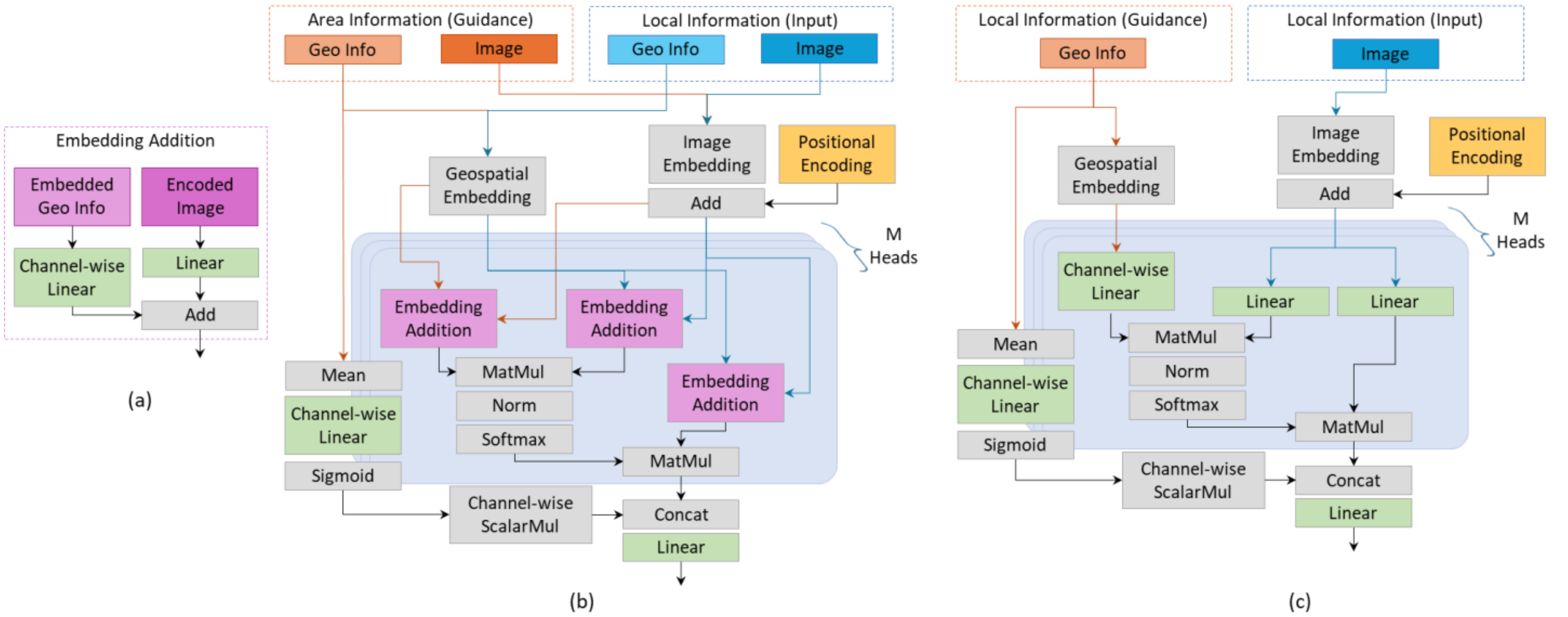

**Fig. 4.** Detailed architectures of the Geo-Context Guided Visual Transformer models: GCGVT-A (b) and GCGVT-G (c), both based on the proposed Guided Attention module. GCGVT-A incorporates local image and geospatial information as input, using the surrounding area as guidance, whereas GCGVT-G integrates image features with geospatial information to analyze how spatial context influences image-based representations.

Unlike approaches that encode geospatial data as a flat tabular vector, a coordinate embedding, or a textual description, the proposed Geospatial Embedding preserves patch-level spatial correspondence between the remote sensing image and auxiliary geospatial layers. This is important because the same aggregate socioeconomic or demographic value may correspond to different visual regions within an image patch, depending on census-tract boundaries and local land-use patterns. By computing area-weighted geospatial values within each image patch and preserving category-specific channels, the embedding provides GA with spatially aligned and semantically decomposed guidance. This design enables the model to condition visual attention on geospatial context while still producing spatial attention maps that can be related back to image regions.

As shown in Fig. 3, geospatial data is extracted based on the spatial boundaries defined by the image patches, ensuring a natural spatial correspondence between the geospatial information and visual content. The value for each variable is calculated as the area-weighted average within this boundary. Each category of geospatial data is maintained as a separate channel throughout both preprocessing and embedding. This category-preserving design supports the head-role constraint in GA by preventing variables from different geospatial categories from being mixed before the role-specific attention and modulation branches. The projection remains learnable within each category, allowing the model to combine related variables flexibly while preserving the semantic boundary needed for interpretation.

This design does not imply that an attention head is determined by one geospatial category alone, because the attention map also depends on the input stream. Rather, the geospatial embedding anchors each head to a category. The geospatial components of the query and the head-modulation weight are derived from the assigned category, while the attention operation still learns interactions between that guidance and the input tokens. As a result, the attention map associated with a head can be interpreted as a category-guided spatial interaction between the input representation and the corresponding geo-contextual factor.

The geospatial embedding is processed in parallel with the image data as it is divided into patches. For each image patch, the corresponding geospatial boundary is overlaid at a specified resolution, and the associated geospatial information is extracted for each segment defined by that boundary. This information may include task-relevant contextual factors such as socio-economic status, education levels, environmental conditions, and other spatially distributed attributes.

For each segment $i$ within a patch, a geospatial factor $f_i$ is weighted by the segment's area $area_i$ relative to the image patch (Fig. 3). The aggregated value of a given geospatial factor $f$ for the entire patch is computed as an area-weighted sum across all segments:

$$f = \frac{\sum_i area_i * f_i}{\sum_i area_i} \tag{10}$$

Each category of geospatial information is then linearly projected to match the dimensionality of its corresponding attention head. The resulting geospatial embedding is subsequently fed into the GA alongside the encoded image features, as illustrated in Fig. 3.

### *D. Geo-Context Guided Visual Transformer Architectures*

The overall architecture of the proposed GCGVT extends transformer-based visual models by replacing the conventional multi-head self-attention mechanism with the proposed GA. This design enables the model to jointly process both the remote sensing imagery and geospatial contextual information through guided multimodal interactions. Based on two complementary hypotheses, we develop two variants of the GCGVT to explore different roles of contextual guidance in multimodal representation learning.

The first variant, GCGVT-Area (GCGVT-A), is motivated by a public health hypothesis that environmental conditions in the broader surroundings may influence local health outcomes. For example, two visually similar residential areas exhibit different respiratory health risks if one is located near industrial zones or along the pathways of wildfire smoke. In this architecture (Fig. 4(b)), the input consists of a high-resolution local RGB image, $X_I^0$, and its corresponding geospatial layers, $X_I^r$ , that capture fine-grained structural and environmental attributes. The guidance is derived from a wider, zoomed-out context that encompasses both imagery $X_G^0$ and geospatial data $X_G^r$, providing broader environmental cues. Through Geospatial Embedding and GA, the model integrates these complementary sources, enabling reasoning about the interactions between local features and regional contextual factors.

The second variant, GCGVT-Geo (GCGVT-G), is inspired by the theory that visual processing is influenced by prior knowledge and contextual cues [111]. The underlying hypothesis is that identical physical environments may be interpreted differently depending on socio-demographic attributes such as household composition or tenancy status. This reasoning parallels how human analysts interpret remote sensing imagery in light of background information. In this architecture (Fig. 4(c)), the input is the local remote sensing image ($X_I^0$ only for (5) and (6)), while the guidance is derived from the geospatial embedding of the same spatial region ($X_G^r$ only for (4)). This configuration represents image interpretation under domain-specific contextual influence. Compared with GCGVT-A, GCGVT-G uses less external context and may therefore yield lower predictive performance. However, its explicit separation between image features (as input) and geospatial contextual information (as guidance) enhances interpretability by clarifying the contribution of each geospatial category.

For both variants, the models are implemented using a ViT-style backbone to isolate the effects of the proposed GA. Nevertheless, the overall architecture is modular and can be adapted to alternative visual transformer backbones, ensuring flexibility for future extensions and broader multimodal applications.

## IV. Experiments

To contextualize our experiments, we briefly restate the task setting in this work. Our objective is to evaluate models on problems in which remote sensing imagery alone may not capture all relevant factors, and where auxiliary geospatial information can provide complementary context. Such tasks, which are common in public health and environmental applications, often involve influences that lie outside the spatial extent or visible content of the image. GA is therefore used to incorporate contextual geospatial layers when the visual input does not fully encode the predictive signal. The following subsections describe the datasets, experimental configurations, and baselines used to assess this capability.

### *A. Dataset*

Existing public datasets and benchmarks predominantly address standard computer vision tasks, such as image classification, semantic segmentation, and object detection. These tasks typically rely on input images that encapsulate all necessary information, rendering them unsuitable for the specific challenges identified in our study. Furthermore, while these datasets are often organized by category or object type, they generally omit geospatial information. To evaluate our model, we developed a dataset that integrates remote sensing as the primary input, augmented with corresponding geospatial information. Health outcomes were chosen as the prediction target, as they can be influenced directly by local features captured in the primary input and indirectly by broader contextual factors in the surrounding area or relevant geospatial information. This dataset fills a gap in the field as a task that can benefit from information beyond the image and a benchmark for multimodal models.

*1) Health Outcomes:* Health outcomes were selected as the target variables for prediction, given their potential to benefit from additional contextual information. We utilized the PLACES dataset [112] from the Centers for Disease Control and Prevention, which offers comprehensive nationwide coverage and high spatial resolution at the census tract level. This dataset provides detailed measurements across various public health dimensions, including health outcomes, prevention efforts, health risk behaviors, disabilities, health status, health-related social needs, and social determinants of health. However, some factors could be directly inferred from economic status or were self-reported by individuals. To ensure reliability, our experiments focused exclusively on health outcomes derived from professional medical evaluations and excluded factors directly reflecting socioeconomic status. Specifically, among 12 total health outcomes, we selected the prevalence of arthritis, cancer, chronic obstructive pulmonary disease, coronary heart disease, depression, diabetes, chronic kidney disease, obesity, and stroke. Current asthma, high cholesterol among screened adults, and all teeth lost among adults older than 65 years were excluded since they could be directly correlated with environmental, socioeconomic, or demographic factors. Data from the year 2020 were used to minimize the impact of the COVID-19 pandemic while preserving the availability of high-resolution imagery. The prevalence values represent model-estimated percentages for each condition within each census tract, ranging from 0 to 100. Since a single sampled image might encompass multiple census tracts, we calculated the area-weighted average for each health outcome to serve as the ground truth for the corresponding sampled image.

*2) Remote Sensing Images:* To minimize the inclusion of unpopulated areas or regions lacking sufficient variation, we used historical redlining zones as a baseline for generating geospatial sampling points. Originally created in the 1930s to guide mortgage lenders, redlining zones categorized residential areas in major U.S. cities based on investment risk. Despite their prohibition decades ago, these zones continue to influence various aspects of residents' lives [113], [114], making them a suitable baseline for sampling images and corresponding health outcomes.

We obtained digitized redlining zone polygons from the Mapping Inequality dataset [115]. This dataset includes redlining designations at four risk levels across 138 cities in 44 U.S. states. By matching these geospatial locations with the PLACES dataset, we identified 14,330 valid areas. The centroid of each area was used as the sampling point for remote sensing images, ensuring minimal overlap between images from adjacent areas.

The remote sensing images were sourced from the National Agriculture Imagery Program (NAIP), which provides high spatial resolution imagery collected during the growing season to ensure consistency across states. Using the centroids of the redlining areas, we sampled images from the year closest to 2020. For each point, two image sizes were generated: local images (640 × 640 pixels) and area images (1280 × 1280 pixels). At a resolution of 60 cm/pixel, these images covered approximately 0.15 and 0.59 square kilometers, respectively. The area images were resized to 640 × 640 pixels using the bicubic interpolation method.

*3) Geospatial Layers:* For geospatial embeddings, we utilized the 2020 American Community Survey (ACS) data [116] provided by the United States Census Bureau. To achieve a balance between high-level insights and sufficient detail, we used the 5-year data profiles, which aggregate geospatial information over a five-year period ending in 2020. The dataset encompasses social, economic, housing, and demographic information for every census tract in the U.S., totaling more than 2,400 layers. However, certain categories were either overly detailed, such as specific construction years of houses or mortgage amounts, or had direct ties to health outcomes like disability rates and health insurance coverage. Such features were excluded to avoid having proxies for the health outcomes.

We selected ten broad categories from the dataset: age and sex, ancestry, education, employment, household, housing, income, origin, race, and residency, resulting in 228 variables. To account for the varying population distributions across states, percentage values were used to normalize the data. These percentages, ranging from 0 to 100, represent the proportion of the population meeting the criteria for each variable. For each image patch, the values were calculated as area-weighted averages derived from all census tracts overlapping with the patch as described in Section III-C.

### B. Experiment Configuration

To evaluate the effectiveness of GCGVT and to characterize the role of each data modality, we designed our experiments around three complementary objectives. First, we assessed the overall multimodal performance of GCGVT when jointly incorporating imagery and geospatial information, enabling direct comparison with remote-sensing vision-language foundation models and graph-based spatial fusion baselines. Second, we examined the influence of geospatial data in isolation and in combination with imagery, motivated by the need to understand how effectively GCGVT leverages auxiliary geospatial layers and how performance changes when such information is limited or partially available. Third, we evaluated the model in an image-only setting to analyze how the GA leverages local and area imagery, and to benchmark against transformer-based architectures and pretrained remote sensing models. This structure ensures that each experiment isolates a specific aspect of the method, providing a comprehensive assessment of its capabilities.

As described in Section III-D, all variants of GCGVT were implemented using a ViT-base-16 backbone with 16×16 pixel image patches. Given the size of the available dataset and the need to isolate the effect of the proposed GA module, we used a compact transformer configuration with four encoder blocks, ten guided-attention heads, token embedding dimension of 128, attention head dimension of 64, and MLP hidden dimension of 1280. The ten attention heads correspond to the ten broad geospatial categories, preserving the category-to-head structure used for interpretability. The concatenated attention-head output was linearly projected back to the 128-dimensional token embedding space. Dropout was set to 0.1 in the attention, output-projection, and feed-forward layers. The class token was removed and replaced with average pooling to facilitate analysis of patch-level interactions. Supplemental Table I provides a comparison of model configurations for different variants. Due to architectural modifications, GCGVT-A and GCGVT-G were initialized with random weights and pretrained on MLRSNet [117] using an image classification task to learn general remote-sensing representations. All datasets were randomly divided using an 8:1:1 train, validation, test split, and models were pretrained and trained for 128 epochs.

For the multimodal evaluation, we compared GCGVT with two groups of baselines. The first group consisted of recent remote-sensing vision-language models: RemoteCLIP (RCLIP) [26], GeoRSCLIP (GeoRS) [27], and SkyCLIP (Sky) [28]. These models used a ViT-large-14 backbone with their publicly released pretrained weights for higher accuracy. Because vision-language models require textual inputs to represent non-visual information, we evaluated two geospatial encoding strategies. The first strategy, denoted T, converted all geospatial variables and their definitions into a natural-language summary generated using the Mistral model [118]. The second strategy, denoted I, represented the same variables as a space-separated sequence of integer values. These encodings allowed vision-language baselines to incorporate structured geospatial data within their predefined text-token constraints.

The second group consisted of graph-based spatial fusion baselines: GATv2 [89] and GraphGPS (GPS) [90]. These baselines were constructed to test whether graph message passing over spatially aligned image-geospatial patches could provide an alternative to the proposed guided-attention fusion. For fair comparison, they used the same local image patches and patch-aligned geospatial embeddings as GCGVT rather than constructing a separate census-tract-level graph. Each patch of the 40×40 patch grid was treated as a graph node, and

TABLE I

OVERALL PERFORMANCE FOR EACH HEALTH OUTCOME WITH MULTIMODAL INPUT

| | **GCGVT-A** | **GCGVT-G** | **GATv2** | **GPS** | **RCLIP-T** | **RCLIP-I** | **GeoRS-T** | **GeoRS-I** | **Sky-T** | **Sky-I** |
|---|---|---|---|---|---|---|---|---|---|---|
| Arthritis | 0.84† | 0.79‡ | 0.77 | 0.79‡ | 0.39 | 0.49 | 0.36 | 0.48 | 0.30 | 0.53 |
| Cancer | 0.86† | 0.85‡ | 0.80 | 0.84 | 0.20 | 0.50 | 0.27 | 0.44 | 0.25 | 0.53 |
| CHD | 0.80† | 0.78‡ | 0.74 | 0.76 | 0.27 | 0.41 | 0.22 | 0.38 | 0.25 | 0.40 |
| COPD | 0.83† | 0.81‡ | 0.79 | 0.81‡ | 0.38 | 0.54 | 0.36 | 0.51 | 0.34 | 0.53 |
| Depression | 0.85† | 0.83‡ | 0.67 | 0.68 | 0.42 | 0.49 | 0.45 | 0.49 | 0.43 | 0.48 |
| Diabetes | 0.88† | 0.88† | 0.86 | 0.87‡ | 0.35 | 0.53 | 0.37 | 0.57 | 0.35 | 0.63 |
| Kidney | 0.84† | 0.82 | 0.82 | 0.83‡ | 0.29 | 0.51 | 0.26 | 0.49 | 0.23 | 0.50 |
| Obesity | 0.90† | 0.89‡ | 0.86 | 0.86 | 0.47 | 0.63 | 0.45 | 0.61 | 0.50 | 0.64 |
| Stroke | 0.85† | 0.84 | 0.84 | 0.85‡ | 0.29 | 0.59 | 0.35 | 0.57 | 0.39 | 0.55 |
| Mean | 0.85†±0.03 | 0.83‡±0.04 | 0.79±0.06 | 0.81±0.06 | 0.34±0.08 | 0.52±0.06 | 0.34±0.08 | 0.50±0.07 | 0.34±0.08 | 0.53±0.07 |
| Param. (M) | 4.13 | 3.47 | 1.86 | 1.13 | 429.20 | 429.20 | 429.20 | 429.20 | 429.20 | 429.20 |

Note: † and ‡ indicate the best and second-best performance, respectively; ties share the same mark. Row Mean shows the mean ± standard deviation across outcomes. Row Param. reports the number of parameters in millions.

the corresponding node feature was formed by concatenating the image patch embedding and the geospatial patch embedding and projecting the result to the graph hidden dimension. Edges were defined using fixed eight-neighbor adjacency on the image-patch grid, including horizontal, vertical, and diagonal neighboring patches. Thus, the graph structure represents local patch-grid adjacency within each image sample, rather than geographic adjacency or connectivity among real-world spatial units. The resulting graph was processed by GATv2 or GraphGPS layers, followed by global mean pooling and the regression head for health-outcome prediction. Both graph baselines used four graph layers and matched the proposed models in directly comparable embedding and regression settings while using graph-specific message-passing layers.

To understand the contribution of geospatial information and the fusion mechanism, we conducted experiments focusing on GCGVT-A under settings where either only geospatial embeddings or both modalities were available. This analysis was motivated by the strong predictive capacity often seen in geospatial datasets and the need to examine how effectively GCGVT can leverage such information when the imagery is less informative or when geospatial variables are limited. By comparing performance across these configurations, we assess both the standalone value of geospatial layers and the added benefit of GCGVT's spatially aligned fusion strategy.

Finally, to evaluate image-only performance and analyze the benefit of integrating multi-resolution visual information, we compared GCGVT-AreaImage (GCGVT-AI) and its local-only counterpart, GCGVT-LocalImage (GCGVT-LI), under conditions where the head weights were disabled. GCGVT-AI maintained its input configuration that used local image as input and the area image as guidance, whereas GCGVT-LI used the local image for both roles, essentially making it a self-attention-based transformer. We benchmarked these models against ViT-base-16 [29], Swin-v2-base [119], BEiT-base [120], PVT-v2-base [121], and the three multimodal baselines. Like GCGVT-AI and GCGVT-LI, these four vision-only models were initialized with random weights and pretrained on MLRSNet for fair comparison. This experiment isolates the contribution of area-level context and evaluates whether GCGVT's architecture can more effectively exploit multi-resolution imagery than standard transformer-based approaches.

## V. RESULTS

### *A. Performance Comparison and Analysis*

Our experiments evaluated the ability of the proposed models to predict multiple health outcomes. These outcomes include prevalence of various diseases at the census tract level. To assess quantitative performance, we measured the coefficient of determination ($R^2$) between model predictions and ground-truth prevalence values. $R^2$ provides a normalized and outcome-independent measure of predictive accuracy, allowing comparison across variables with differing means and variances. The following subsections report results for the three experimental settings introduced previously, corresponding directly to the evaluation design described in Section IV-B.

*1) Overall Performance:* To assess multimodal predictive performance, we compared GCGVT-A and GCGVT-G with recent remote-sensing vision-language foundation models and graph-based spatial fusion models using both imagery and geospatial variables (Table I). The vision-language baselines used the two geospatial encoding strategies described in Section IV-B, whereas the graph-based models provide a complementary comparison by directly incorporating geospatial embeddings into spatially structured representations. This experiment addresses our first objective: evaluating whether the proposed method improves multimodal geospatial fusion relative to both language-aligned pretrained models and graph-based spatial modeling approaches.

As shown in Table I, GCGVT-A achieved the highest overall performance, with a mean $R^2$ of 0.85, followed by GCGVT-G with a mean $R^2$ of 0.83. The graph-based models were substantially stronger than the vision-language baselines, with GPS and GATv2 achieving mean $R^2$ values of 0.81 and 0.79, respectively. This difference is expected because graph-based models can incorporate structured geospatial variables more directly and can model spatial relationships among neighboring patches. In contrast, the vision-language baselines were originally developed for image-text alignment and require structured geospatial variables to be represented through indirect textual or token-based encodings. Nevertheless, both graph-based models remained below GCGVT-A and GCGVT-G, suggesting that patch-grid message passing alone does not fully replace the proposed guidance-based mechanism for multimodal geospatial prediction.

The outcome-level differences further clarify this pattern. For outcomes such as diabetes and stroke, the graph-based models approached the performance of GCGVT, suggesting that local spatial adjacency and structured geospatial variables provide strong predictive information for these conditions. In contrast, the proposed models showed a clearer advantage for depression. This may indicate that depression prevalence is less directly captured by the included census-derived variables alone and may depend more strongly on visual cues related to living environment or other contextual features observable in imagery. In this setting, the asymmetric input-guidance formulation may be beneficial because it allows geospatial context to modulate the interpretation of image features rather than merging all information into a single node representation. These outcome-specific patterns motivate both the statistical reliability analysis below and the ablation and visualization analyses that follow.

To quantify the reliability of the improvements over the closest non-GCGVT baseline, we performed paired bootstrap testing against GraphGPS using the same held-out test samples. For each outcome, we generated 10,000 bootstrap resamples of the test set. In each resample, $R^2$ was recomputed for both models, and the paired difference was calculated as $R^2$ of GCGVT models minus $R^2$ of GraphGPS. The 95% confidence intervals were obtained using the percentile bootstrap, and two-sided bootstrap p-values were adjusted using the Holm procedure across the reported paired comparisons. As summarized in Supplemental Table II, GCGVT-A and GCGVT-G showed statistically significant gains for five and two outcomes after Holm correction, respectively. These results provide statistical support for the performance gains in Table I.

Table I also reports the number of trainable parameters for each model. The CLIP-based remote-sensing baselines contain substantially larger parameter counts, yet their predictive performance remains lower than that of the proposed models, suggesting that larger pretrained vision-language encoders do not directly translate to better use of structured geospatial context. The graph-based fusion models are parameter-efficient and achieve competitive performance, but GCGVT-A obtains the highest mean $R^2$, while GCGVT-G achieves a favorable balance between accuracy, compactness, and interpretability. These results indicate that the improvement of GCGVT is not due to model size, but to the spatially aligned geospatial embedding and guided-attention mechanism.

To further examine whether the performance in Table I depends strongly on design-specific hyperparameters in GA, we conducted additional sensitivity analyses for both GCGVT-A and GCGVT-G. We evaluated variants with category-mixed guidance, five random geospatial category groupings, and five hidden-dimension settings for the head-modulation network. Compared with the original head-category mapping, the mixed and random category variants produced only small performance differences, while weakening the direct correspondence between geospatial categories and attention heads. Varying the hidden dimension of the head-modulation network also had negligible effect. Across the tested settings, the mean change in $R^2$ remained small, approximately -0.01 to 0.00 for both models. These results indicate that instead of a performance-optimal configuration, the semantic one-to-one mapping is an interpretation-oriented structural constraint that preserves category-level traceability while maintaining competitive predictive performance. Detailed results and configuration notes are provided in Supplemental Table III.

TABLE II
MEAN PERFORMANCE OF GCGVT-A WITH LIMITED GEOSPATIAL INFORMATION

| Category | Image & Geo | Geo Only | Diff |
|---|---|---|---|
| Age | 0.64±0.09 | 0.36±0.17 | 0.28±0.13 |
| Ancestry | 0.60±0.12 | 0.39±0.12 | 0.21±0.06 |
| Education | 0.61±0.09 | 0.44±0.13 | 0.17±0.11 |
| Employment | 0.69±0.06 | 0.51±0.16 | 0.19±0.13 |
| Household | 0.68±0.05 | 0.53±0.12 | 0.16±0.12 |
| Housing | 0.70±0.05 | 0.63±0.10 | 0.07±0.10 |
| Income | 0.75±0.04 | 0.66±0.13 | 0.09±0.11 |
| Origin | 0.48±0.12 | 0.30±0.08 | 0.18±0.10 |
| Race | 0.60±0.13 | 0.42±0.16 | 0.18±0.13 |
| Residency | 0.46±0.10 | 0.09±0.08 | 0.41±0.12 |
| All | 0.85±0.03 | 0.83±0.04 | 0.02±0.03 |

Note: Column Image & Geo shows the mean ± standard deviation across all health outcomes using both image and geospatial information, and Geo Only uses only the geospatial embeddings. Column Diff shows the mean ± standard deviation of the difference between Image & Geo and Geo Only across outcomes.

*2) Ablation Study of Geospatial Information:* To isolate the contribution of structured geospatial data and examine the role of remote sensing imagery, we evaluated the GCGVT-A architecture using only geospatial embeddings as input. This setting examines how effectively the model leverages nonvisual information and how performance changes when imagery is removed. We further assessed the predictive value of each geospatial category individually. These evaluations correspond to our second objective of characterizing the standalone and partial utility of geospatial variables.

Table II shows the mean performance of GCGVT-A across all health outcomes with limited geospatial information. Detailed results for each outcome can be found in Supplemental Table IV. When all geospatial categories were used without imagery (row All), GCGVT-A achieved performance close to the full image-and-geospatial setting, with mean $R^2$ values of 0.83±0.04 and 0.85±0.03, respectively. The mean difference of

TABLE III
IMAGE-ONLY PERFORMANCE COMPARISON

| | **GCGVT-AI** | **GCGVT-LI** | **RCLIP** | **GeoRS** | **Sky** | **ViT** | **Swin-v2** | **BEiT** | **PVTv2** |
|---|---|---|---|---|---|---|---|---|---|
| Arthritis | 0.42† | 0.39‡ | 0.38 | 0.39‡ | 0.36 | 0.25 | 0.39‡ | 0.29 | 0.32 |
| Cancer | 0.26 | 0.19 | 0.32‡ | 0.33† | 0.31 | 0.26 | 0.27 | 0.18 | 0.26 |
| CHD | 0.28‡ | 0.26 | 0.27 | 0.18 | 0.29† | 0.18 | 0.24 | 0.22 | 0.21 |
| COPD | 0.44† | 0.39‡ | 0.31 | 0.39‡ | 0.34 | 0.30 | 0.31 | 0.19 | 0.31 |
| Depression | 0.61† | 0.58‡ | 0.41 | 0.43 | 0.42 | 0.46 | 0.40 | 0.42 | 0.36 |
| Diabetes | 0.42‡ | 0.39 | 0.41 | 0.44† | 0.40 | 0.20 | 0.28 | 0.24 | 0.27 |
| Kidney | 0.31† | 0.27 | 0.27 | 0.26 | 0.28‡ | 0.15 | 0.16 | 0.12 | 0.15 |
| Obesity | 0.55‡ | 0.51 | 0.55‡ | 0.58† | 0.55‡ | 0.32 | 0.40 | 0.37 | 0.40 |
| Stroke | 0.35 | 0.31 | 0.37† | 0.37† | 0.36† | 0.18 | 0.20 | 0.14 | 0.24 |
| Mean | 0.41†±0.11 | 0.37‡±0.12 | 0.37‡±0.08 | 0.37‡±0.11 | 0.37‡±0.08 | 0.25±0.09 | 0.29±0.08 | 0.24±0.09 | 0.28±0.07 |

Note: † and ‡ mark the best and second-best performance, respectively, for the outcome. Row Mean shows the mean ± standard deviation for the model.

0.02±0.03 indicates that when the complete geospatial variable set is available, the average incremental predictive contribution of remote sensing imagery is modest. This result suggests strong standalone associations between the geospatial variables and health outcomes.

This pattern was also observed in the graph-based baselines. When evaluated using only patch-aligned geospatial variables, without image embeddings, GATv2 and GraphGPS achieved mean $R^2$ values of 0.80±0.06 and 0.81±0.06, respectively. These results further support the strong predictive value of the geospatial layers and suggest that much of the predictive power of graph-based models might come from spatially structured geospatial information. Compared with these geo-only results, GCGVT-A still showed a modest overall improvement when imagery was included, suggesting that the proposed guided-attention formulation can incorporate visual information for multimodal prediction, although the average gain is limited when comprehensive geospatial variables are already available.

The role of imagery becomes clearer when geospatial information is limited. Using only one geospatial category led to a substantial drop in performance in almost all cases compared with using imagery together with that category. Among the individual categories, economic-status-related factors, including housing and income, were most strongly associated with various health outcomes. Adding remote sensing imagery to a single geospatial category improved prediction by a mean difference of 0.07 to 0.41 across categories, suggesting the effectiveness of the combination of remote sensing imagery with sparse geospatial information. The table also revealed differences across outcomes. For depression, no single geospatial category achieved performance comparable to that of other outcomes (Supplemental Table IV), suggesting that visual features related to the residential environment may provide useful contextual information and play a more important role for this outcome.

These results suggest that remote sensing imagery can be especially valuable in reduced-information settings, where complete geospatial data are missing, unreliable, or difficult to obtain. We further evaluated two-category combinations alongside remote sensing imagery and compared them with the baseline that uses all geospatial categories without imagery (G with All categories in Supplemental Table IV). The top three performing combinations are reported in Supplemental Table V. Across all health outcomes, prediction was feasible with only two geospatial categories in addition to imagery, with the best-performing combinations remaining close to the full geospatial baseline. In some cases, such as AgeSex+Housing for arthritis and Income+Race for depression, the reduced-input models even outperformed the full geospatial-only model. These results highlight a practical advantage that competitive prediction is possible with substantially less geospatial information, potentially reducing reliance on costly or difficult-to-obtain geospatial variables.

*3) Ablation Study of Remote Sensing Imagery:* To quantify the benefit of incorporating area-level visual context, we also performed image-only experiments. In this setting, GCGVT-AI used local imagery with area-level guidance, while GCGVT-LI provided a controlled baseline using only local imagery. Benchmark transformer models received the area image as their sole input. This analysis addresses our third objective: determining the effectiveness of GCGVT's GA module for multi-resolution visual information in the absence of geospatial data.

As shown in Table III, GCGVT-AI consistently outperformed GCGVT-LI across all outcomes, demonstrating that incorporating area-level guidance improves prediction. Moreover, GCGVT-AI achieved higher overall performance than other models that also used area images as input, indicating that the guidance mechanism effectively integrates fine-grained local details with broader contextual information. This ability to fuse information across resolutions enhances performance in typical computer vision tasks and highlights the value of using complementary images to capture complex geospatial relationships. The improved performance achieved by fusing local and area images demonstrates not only the benefit of incorporating related imagery at different spatial resolutions but also the potential to effectively integrate images with distinct yet intrinsically related visual features.

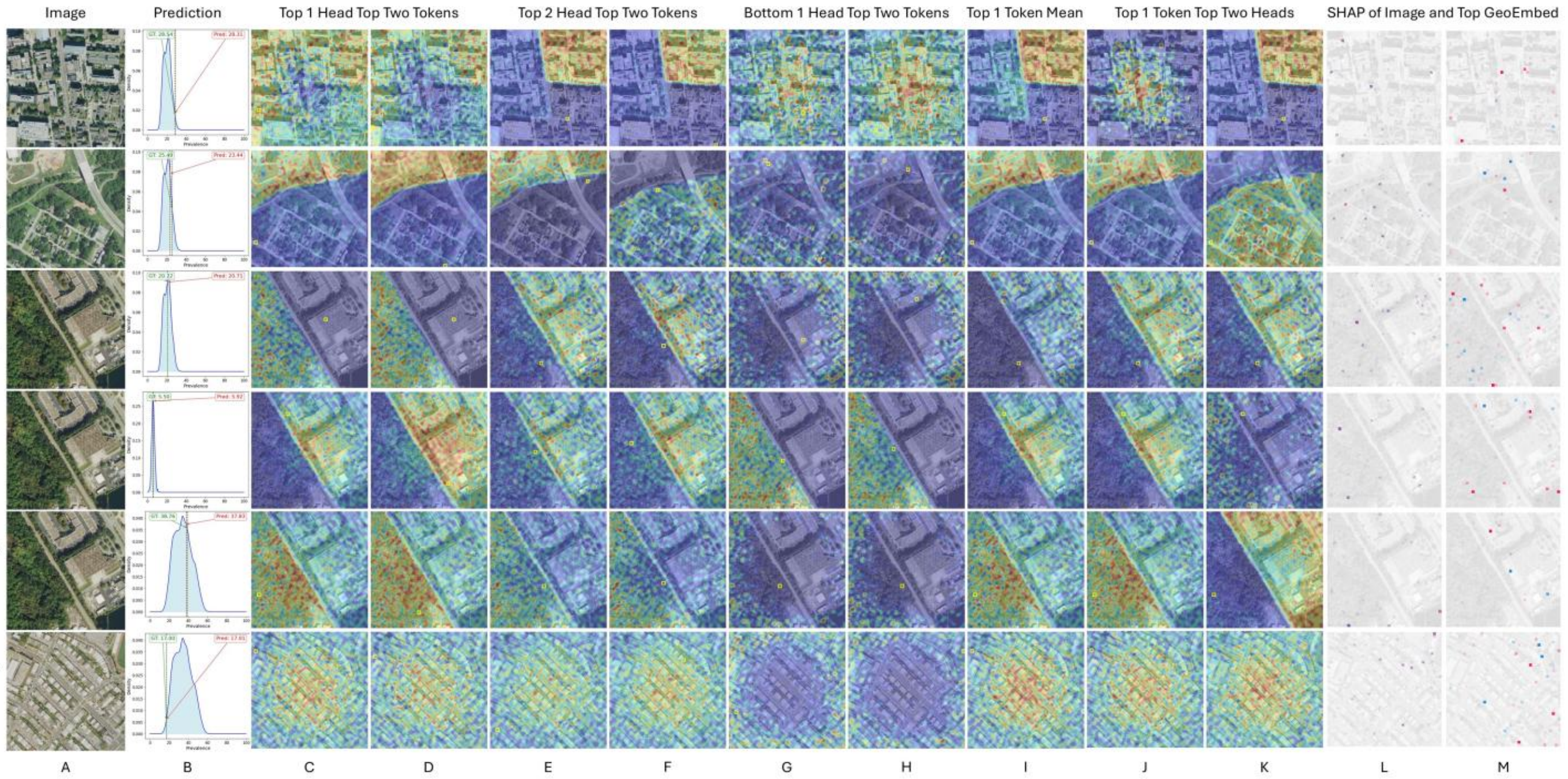

**Fig. 5.** Examples of attention map visualization of various outcomes by GCGVT-G. Each row represents one input sample. Column A shows the raw input image; Column B overlays the model prediction against ground truth on a kernel density estimate of all ground truth values. Columns C–D display the top two tokens from the category with highest head weight; Columns E–F from the second-highest category; and Columns G–H from the lowest-ranked category. Column I shows the most influential token averaged across heads, and columns J–K present the top two heads associated with this token. Columns L–M visualize SHAP value for the input image and the geospatial embedding of the top category. Yellow squares in columns C to K are tokens of interest.

### *B. Prediction Visualization and Interpretation*

With the proposed GA, the generated attention weights are structured to support category-level interpretation. Because unconstrained attention maps alone may not constitute faithful explanations [106], we do not interpret attention weights in isolation. Instead, we interpret them jointly with the category-to-head constraint and the guidance-derived head weights. The head modulation mechanism provides a structured way to examine the relative importance assigned to different geospatial categories, while the corresponding attention maps indicate which regions are emphasized under each category-specific guidance signal. Under the category-to-head constraint, the guidance-derived head weights are interpreted as learned soft relevance scores for geospatial categories, and the corresponding attention maps are interpreted as category-conditioned spatial emphasis patterns. Thus, the visualizations are used to characterize structured image–geospatial interactions under category-specific guidance, rather than as standalone tests of explanation faithfulness or causal attribution.

We further report quantitative descriptors for the representative cases shown in Fig. 5 and Fig. 6. Following the notation in Section III-B, the head weight $h_i$ is used as the category-level relevance score for the geospatial category assigned to head $i$. For a visualized attention map, we normalize the attention values over image tokens and compute a normalized entropy-based concentration score,

$$C_{\text{attn}} = 1 - \frac{-\sum_{n=1}^{N} a_n \log(a_n)}{\log N}, \quad (11)$$

where $a_n$ is the normalized attention value of token $n$, and $N$ is the number of image tokens. A higher value indicates that the attention map is more spatially concentrated, whereas a lower value indicates a more diffuse map. To compare whether two categories or outcomes emphasize different image regions, we also compute the cosine distance between their normalized attention maps,

$$D_{\text{attn}}(a, b) = 1 - \frac{a^{\top} b}{\|a\| \|b\|}. \quad (12)$$

For same-image comparisons across different outcomes, we additionally report the Spearman correlation between the corresponding head-weight vectors to quantify whether the category rankings change. These measures are not intended as standalone faithfulness metrics; rather, they provide numerical support for the case-level interpretation of category relevance, spatial concentration, and differences among category-guided attention patterns.

*1) Visualization of GCGVT-G:* We first analyze GCGVT-G visualizations because this variant employs separate pathways for image input and geospatial guidance, making the category-guided attention maps clearer to interpret than those of GCGVT-A. Attention maps were generated in two ways: (1) by identifying the most important head based on head weights and visualizing the tokens with the highest summed weights, and (2) by identifying the most important token in the guidance via head-averaged weights and visualizing both its averaged attention map and the most influential heads. These complementary approaches provide different views of the learned image-geospatial interactions.

Fig. 5 presents representative examples of GCGVT-G visualizations. Each row corresponds to a specific input and health outcome. For reference, the numbers in parentheses indicate the learned head weights, which represent the relative importance assigned to each geospatial category in GA. The first three rows show predictions of depression prevalence. The first example depicts a moderate-density residential area, where the most influential categories are origin ($h$=0.87) and employment ($h$=0.87). Their corresponding attention maps highlight different spatial regions: origin guidance emphasizes the upper portion of the image, whereas employment guidance concentrates on the top-left residential block. This separation is also reflected quantitatively with $D_{\mathrm{attn}}$=0.61. In contrast, the least influential category, age, has a nearly zero head weight and shows a much larger distance from the origin-guided map ($D_{\mathrm{attn}}$=0.87), supporting its limited relevance in this example.

The second and third rows show areas with greater vegetation and lower depression prevalence. In both cases, education ($h$=0.99 and 0.77) and origin ($h$=0.50 and 0.57) are among the most influential categories, yet the associated image features differ. For education, attention in row 2 centers on farmland rather than residential zones, while in row 3 it highlights a natural area adjacent to a residential complex. For origin, row 2 emphasizes both farmland and low-density housing, whereas row 3 focuses more on the residential complex. In the third row, the education- and origin-guided maps have $D_{\mathrm{attn}}$=0.67, indicating that the two categories guide the model toward distinct spatial evidence even within the same image. These findings suggest that different geospatial categories can modulate the same visual scene through separate image-token interactions.

Across these depression examples, the model achieves high predictive accuracy while producing category-conditioned attention patterns that differentiate how geospatial categories interact with visual tokens. Because GA constrains each head to a specific geospatial category, the resulting maps allow salient spatial regions to be examined in relation to semantically meaningful population and contextual factors. For example, employment-related guidance emphasizes built-environment structures such as housing blocks or road-access patterns, whereas origin-related guidance may emphasize non-residential or mixed land-use regions. These patterns should be interpreted as category-conditioned visual cues learned by the model rather than as direct causal attribution of the prediction. In contrast, SHAP visualizations highlight important pixels but do not distinguish which geospatial category they relate to. The GCGVT-G maps therefore provide a more structured category-conditioned interpretation of cross-modal interactions.

The third to fifth rows of Fig. 5 further illustrate outcome-specific interpretation using the same input image for depression, cancer, and obesity prediction. For depression, the two most influential categories are education ($h$=0.77) and origin ($h$=0.57). For cancer, origin remains dominant ($h$=0.96), but household becomes the second most influential category ($h$=0.55). For obesity, origin is again dominant ($h$=0.97), while ancestry becomes the second category ($h$=0.46). The head-weight rankings differ across outcomes, with Spearman correlations of 0.78 between depression and cancer, 0.37 between depression and obesity, and 0.58 between cancer and obesity. The category-specific attention maps are also separated within each outcome, with $D_{\mathrm{attn}}$ values of 0.67 for depression, 0.82 for cancer, and 0.42 for obesity between the top two categories. These descriptors provide quantitative support for the visual observation that the same image can produce different category-guided attention patterns depending on the predicted health outcome.

Finally, some less informative visualizations were observed. The last row of Fig. 5 shows a densely built residential area with little spatial variation. Although the prediction remains accurate, the top-ranked categories, origin, housing, and income, produce highly diffuse maps, with low attention concentration scores of $C_{\mathrm{attn}}$=0.02, 0.03, and 0.01, respectively. Their attention maps are also relatively similar, with distances of only 0.19 to 0.20 from the top-ranked origin map. This is consistent with the qualitative observation that attention maps across categories converge on central housing regions, resulting in limited spatial specificity when the image lacks meaningful visual heterogeneity.

*2) Visualization of GCGVT-A:* For GCGVT-A, both the input and the guidance consist of imagery paired with geospatial embeddings. Consequently, the learned interactions occur not only between geospatial categories and local image tokens, but also between local and broader contextual regions. This design improves predictive performance by incorporating wider spatial context, but it also makes the resulting attention maps less directly separable than those of GCGVT-G because visual and geospatial information are embedded in both streams. Therefore, the GCGVT-A visualizations are interpreted primarily as region-level interaction patterns, while GCGVT-G provides more direct category-to-image interpretation. Fig. 6 presents representative visualization examples of GCGVT-A using the same layout as Fig. 5.

Although the maps are generally more complex, region-specific interactions remain evident in several cases. In the obesity example, the housing category receives a high head weight of 0.94 and produces a spatially concentrated map over the apartment-complex region ($C_{\mathrm{attn}}$=0.16). This map is also strongly separated from the origin-guided map ($D_{\mathrm{attn}}$=0.97), indicating that the housing head captures a distinct spatial interaction rather than simply repeating the dominant category. This pattern is consistent with the visual observation that GCGVT-A can associate broader contextual guidance with local housing structures.

Rows corresponding to arthritis and diabetes further illustrate outcome-specific behavior within the same region. For arthritis, the most influential categories are origin ($h$=1.00), income ($h$=0.83), and housing ($h$=0.64), and the origin- and income-guided maps are both spatially concentrated ($C_{\mathrm{attn}}$=0.34 and 0.30) while remaining highly distinct ($D_{\mathrm{attn}}$=0.94). For diabetes, the dominant categories shift to employment ($h$=1.00), housing ($h$=1.00), and education ($h$=0.73), with lower concentration for the top two maps ($C_{\text{attn}}$=0.17 and 0.14). The head-weight rankings between the arthritis and diabetes predictions have a Spearman correlation of only 0.04, indicating that the model assigns substantially different category-guided interactions patterns to different outcomes in the same spatial region.

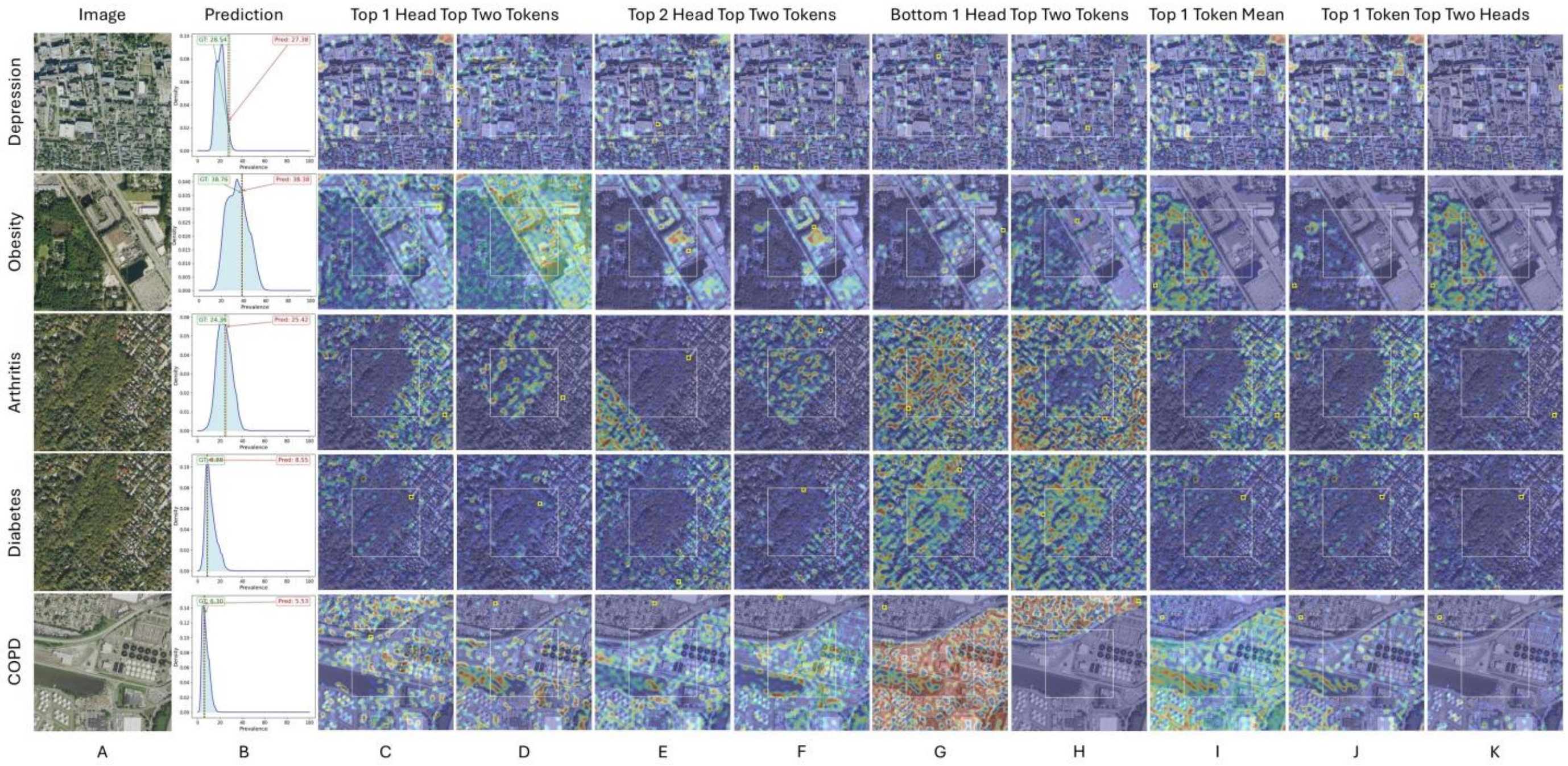

**Fig. 6.** Examples of attention map visualization of various outcomes by GCGVT-A. Similar to Fig. 5, each row represents one input sample. Column A shows the raw input image; Column B overlays the model prediction against ground truth on a kernel density estimate of all ground truth values. Columns C–D display the top two tokens from the category with highest head weight; Columns E–F from the second-highest category; and Columns G–H from the lowest-ranked category. Column I shows the most influential token averaged across heads, and columns J–K present the top two heads associated with this token. Yellow squares in columns C to K are tokens of interest.

The COPD example further illustrates the role of broader context. The image contains a residential area adjacent to an industrial zone, and the most influential categories include employment ($h$=1.00), origin ($h$=0.99), housing ($h$=0.88), and income ($h$=0.79). The employment- and origin-guided maps are both spatially focused ($C_{\text{attn}}$=0.54 and 0.37) and highly separated from each other ($D_{\text{attn}}$=0.99), suggesting that the learned interactions distinguish multiple regional cues relevant to the prediction. This pattern is consistent with the qualitative observation that GCGVT-A can attend to contextual areas outside the local residential core when predicting lung-related outcomes.

Overall, these examples indicate that GCGVT-A can characterize meaningful region-level interactions, particularly when broader surrounding context is relevant. However, because image and geospatial information are embedded in both the input and guidance streams, its attention-based visualizations should be interpreted as region-level interaction patterns rather than as clean category-to-image attribution or standalone faithfulness evidence. In this respect, GCGVT-A provides stronger contextual modeling, whereas GCGVT-G offers clearer category-specific visual interpretation.

### C. Summary

Across all evaluation tasks, GCGVT demonstrates an advantage over vision-only baselines, remote-sensing vision-language models, and graph-based spatial fusion baselines. These results highlight the effectiveness of incorporating spatially aligned geospatial information through GA, enabling the model to leverage complementary cues that are not directly observable in remote sensing imagery. Between the two variants, GCGVT-A benefits from broader contextual embeddings and achieves the highest overall performance, whereas GCGVT-G provides more interpretable attention maps by restricting guidance to geospatial layers while using imagery as the sole primary input.

The models also exhibit practical value in reduced-information settings. Competitive performance can be achieved with only a small subset of geospatial variables when remote sensing imagery is included, indicating that the framework may remain effective when comprehensive census-scale inputs are unavailable or costly to obtain. Beyond predictive accuracy, the case-level visualization analysis shows that category-specific head weights and attention maps provide a structured way to examine image–geospatial interactions. The attention concentration scores and inter-map distances support the qualitative observations in Fig. 5 and Fig. 6, indicating that different geospatial categories and outcome-specific prediction tasks are associated with distinct spatial emphasis patterns in the learned model. These results suggest that GCGVT provides improved prediction performance together with interpretable cross-modal cues relevant to public-health analysis.

## VI. Conclusion

This study presented the Geo-Context Guided Visual Transformer, a multimodal framework for incorporating structured geospatial context into transformer-based remote sensing analysis. The proposed design addresses a setting in which relevant contextual factors may be spatially distributed, indirectly related to visual appearance, or not directly visible in the image. Through patch-aligned Geospatial Embeddings and Guided Attention, GCGVT uses auxiliary geospatial information to condition visual feature processing while maintaining a distinction between the primary input and the guidance signal. This asymmetric design differentiates the framework from generic image-text alignment, conventional visual transformers, and fusion strategies that mix modalities without preserving interpretable guidance pathways.

Experimental results on health-outcome prediction showed that GCGVT improves performance over vision-only models, pretrained remote-sensing vision-language models, and patch-grid graph fusion baselines. The two variants highlight complementary strengths: GCGVT-A benefits from broader contextual information and achieves stronger overall prediction, whereas GCGVT-G provides cleaner category-specific attention patterns by using geospatial variables as guidance for image-based prediction. The ablation studies further indicate that imagery is especially useful when only partial geospatial information is available, suggesting practical value in settings where comprehensive GIS or census-derived variables are incomplete, costly, or difficult to collect. Beyond prediction, the learned head weights and attention maps provide structured visual cues for examining how different geospatial categories relate to spatial emphasis patterns, offering a useful starting point for subsequent public-health interpretation.

Several limitations remain. When the full set of geospatial variables is available, the standalone geospatial setting achieves performance close to the full multimodal model, indicating that the average incremental contribution of imagery should be interpreted cautiously. In addition, the one-to-one category-to-head assignment is intended as an interpretability-oriented constraint rather than a universally optimal fusion design, and the two-layer scoring network serves as a compact nonlinear gating function rather than an exhaustive design choice. The attention-based visualizations should also be viewed as descriptive analyses of learned image-geospatial interaction patterns, not as direct evidence of causal attribution or explanation faithfulness. Future work may explore more flexible category-to-head mappings, alternative guidance-scoring functions, scalable geospatial representations, temporal extensions, and applications beyond health-outcome prediction.

Overall, GCGVT provides a flexible foundation for geospatially guided multimodal remote sensing, supporting both predictive modeling under varying levels of contextual data availability and structured analysis of cross-modal spatial interactions.

## Acknowledgment

The Mistral model [118] was used to create natural-language summaries of geospatial information for the compared "-T" vision-language baseline models in Table I. Examples of generated summaries are available in Supplemental Table VI.

# Supplemental Document for Observing Health Outcomes Using Remote Sensing Imagery and Geo-Context Guided Visual Transformer

Yu Li, *Member, IEEE*, Guilherme N. DeSouza, *Senior Member, IEEE*, Praveen Rao, *Senior Member, IEEE*, Chi-Ren Shyu, *Senior Member, IEEE*

SUPPLEMENTAL TABLE I
DETAILED MODEL CONFIGURATIONS FOR THE GCGVT VARIANTS

| Settings | Multimodal Benchmark | | Image-Only Benchmark | |
|---|---|---|---|---|
| Model | GCGVT-A | GCGVT-G | GCGVT-AI | GCGVT-LI |
| Input stream | local imagery + local geospatial embedding | local imagery | local imagery | local imagery |
| Guidance stream | area imagery + area geospatial embedding | local geospatial embedding | area imagery | local imagery |
| Image input embedding | enabled | enabled | enabled | enabled |
| Geospatial input embedding | enabled | disabled | disabled | disabled |
| Image guidance embedding | enabled | disabled | enabled | enabled |
| Geospatial guidance embedding | enabled | enabled | disabled | disabled |
| Head modulation | enabled | enabled | disabled | disabled |
| Number of heads | 10 | 10 | 10 | 10 |
| Number of transformer layers | 4 | 4 | 4 | 4 |
| Token embedding dimension | 128 | 128 | 128 | 128 |
| MLP hidden dimension | 1280 | 1280 | 1280 | 1280 |
| Attention head dimension | 64 | 64 | 64 | 64 |
| Pooling | mean | mean | mean | mean |
| Dropout | 0.1 | 0.1 | 0.1 | 0.1 |

SUPPLEMENTAL TABLE II
DETAILED STATISTICAL TESTS FOR PERFORMANCE DIFFERENCE BETWEEN GCGVT MODELS AND GRAPHGPS

| | Performance | | | GCGVT-A vs. GPS | | | GCGVT-G vs. GPS | | |
|---|---|---|---|---|---|---|---|---|---|
| Outcome | GCGVT-A $R^2$ | GCGVT-G $R^2$ | GPS $R^2$ | $\Delta R^2$ | p | $p_{Holm}$ | $\Delta R^2$ | p | $p_{Holm}$ |
| Arthritis | 0.84 [0.82, 0.85] | 0.79 [0.76, 0.81] | 0.79 [0.77, 0.82] | 0.04 [0.03, 0.06] | < 0.001 | 0.004 | -0.00 [-0.02, 0.01] | 0.363 | 1.000 |
| Cancer | 0.86 [0.85, 0.88] | 0.85 [0.83, 0.86] | 0.84 [0.83, 0.86] | 0.02 [0.01, 0.03] | < 0.001 | 0.004 | 0.00 [-0.01, 0.02] | 0.573 | 1.000 |
| CHD | 0.80 [0.77, 0.83] | 0.78 [0.75, 0.81] | 0.76 [0.73, 0.78] | 0.04 [0.02, 0.06] | 0.001 | 0.010 | 0.03 [0.00, 0.05] | 0.029 | 0.321 |
| COPD | 0.83 [0.80, 0.86] | 0.81 [0.78, 0.84] | 0.81 [0.79, 0.84] | 0.02 [-0.00, 0.03] | 0.109 | 1.000 | -0.00 [-0.02, 0.01] | 0.687 | 1.000 |
| Depression | 0.85 [0.83, 0.87] | 0.83 [0.81, 0.85] | 0.68 [0.64, 0.72] | 0.17 [0.14, 0.20] | 0.000 | 0.004 | 0.15 [0.12, 0.19] | < 0.001 | 0.004 |
| Diabetes | 0.88 [0.86, 0.90] | 0.88 [0.86, 0.89] | 0.87 [0.85, 0.89] | 0.01 [-0.01, 0.02] | 0.214 | 1.000 | 0.00 [-0.01, 0.02] | 0.408 | 1.000 |
| Kidney | 0.84 [0.81, 0.86] | 0.82 [0.79, 0.84] | 0.83 [0.81, 0.85] | 0.01 [-0.01, 0.03] | 0.318 | 1.000 | -0.01 [-0.03, 0.00] | 0.126 | 1.000 |
| Obesity | 0.90 [0.89, 0.91] | 0.89 [0.87, 0.90] | 0.86 [0.85, 0.88] | 0.04 [0.03, 0.05] | < 0.001 | 0.004 | 0.03 [0.02, 0.04] | < 0.001 | 0.004 |
| Stroke | 0.85 [0.82, 0.88] | 0.84 [0.81, 0.86] | 0.85 [0.82, 0.87] | 0.00 [-0.03, 0.02] | 0.757 | 1.000 | -0.01 [-0.03, 0.01] | 0.369 | 1.000 |

Note: Values were computed on the held-out test set using 10,000 bootstrap resamples. $R^2$ [95% CI] reports the percentile bootstrap confidence interval. $\Delta R^2$ [95% CI] reports the $R^2$ of GCGVT models minus $R^2$ of GraphGPS. Paired bootstrap comparisons used identical resampled test indices for both models. Column p is the two-sided bootstrap p-value for $H_0$: $\Delta R^2 = 0$, and column $p_{Holm}$ is Holm-adjusted across all reported paired comparisons.

SUPPLEMENTAL TABLE III
SENSITIVITY ANALYSIS OF GUIDED ATTENTION CONFIGURATIONS

| | GCGVT-A | | | GCGVT-G | | |
|---|---|---|---|---|---|---|
| $R^2$ | Mixed | Random | GateConfig | Mixed | Random | GateConfig |
| Arthritis | 0.83 | 0.81 | 0.83 | 0.81 | 0.79 | 0.80 |
| Cancer | 0.84 | 0.84 | 0.86 | 0.84 | 0.83 | 0.82 |
| CHD | 0.80 | 0.79 | 0.79 | 0.77 | 0.77 | 0.78 |
| COPD | 0.85 | 0.83 | 0.83 | 0.83 | 0.82 | 0.81 |
| Depression | 0.80 | 0.82 | 0.84 | 0.81 | 0.76 | 0.82 |
| Diabetes | 0.89 | 0.88 | 0.89 | 0.87 | 0.87 | 0.88 |
| Kidney | 0.84 | 0.83 | 0.84 | 0.83 | 0.81 | 0.82 |
| Obesity | 0.89 | 0.89 | 0.90 | 0.87 | 0.87 | 0.89 |
| Stroke | 0.86 | 0.85 | 0.86 | 0.84 | 0.83 | 0.84 |
| Mean | 0.84±0.03 | 0.84±0.03 | 0.85±0.03 | 0.83±0.03 | 0.82±0.04 | 0.82±0.04 |

| | GCGVT-A | | | GCGVT-G | | |
|---|---|---|---|---|---|---|
| $\Delta R^2$ | Mixed | Random | GateConfig | Mixed | Random | GateConfig |
| Arthritis | -0.01 | -0.03 | 0.00 | 0.02 | 0.01 | 0.02 |
| Cancer | -0.02 | -0.02 | -0.01 | 0.00 | -0.01 | -0.03 |
| CHD | 0.00 | -0.01 | -0.02 | -0.02 | -0.01 | -0.01 |
| COPD | 0.01 | -0.01 | -0.01 | 0.01 | 0.00 | 0.00 |
| Depression | -0.05 | -0.03 | -0.01 | -0.02 | -0.07 | -0.02 |
| Diabetes | 0.01 | 0.00 | 0.01 | -0.01 | -0.01 | 0.00 |
| Kidney | 0.00 | -0.01 | 0.01 | 0.01 | 0.00 | 0.00 |
| Obesity | -0.01 | -0.01 | 0.00 | -0.02 | -0.02 | 0.00 |
| Stroke | 0.01 | 0.00 | 0.01 | 0.00 | 0.00 | 0.00 |
| Mean | -0.01±0.02 | -0.01±0.01 | 0.00±0.01 | 0.00±0.02 | -0.01±0.02 | 0.00±0.01 |

Note: $\Delta R^2$ is computed relative to the default semantic one-to-one category-to-head GA configuration. Column Mixed uses category-mixed guidance, Random averages five random geospatial variable groupings, and GateConfig varies the dimensions of the head-modulation network across five settings, including hidden dimension sizes 32, 64, 128, 32-16, and 64-32. The Mean row reports mean ± standard deviation across outcomes.

SUPPLEMENTAL TABLE IV
DETAILED PERFORMANCE OF GCGVT-A WITH LIMITED GEOSPATIAL INFORMATION

| Category | Arthritis | | Cancer | | CHD | | COPD | | Depression | | Diabetes | | Kidney | | Obesity | | Stroke | |
|---|---|---|---|---|---|---|---|---|---|---|---|---|---|---|---|---|---|---|
| | I+G | G | I+G | G | I+G | G | I+G | G | I+G | G | I+G | G | I+G | G | I+G | G | I+G | G |
| Age | 0.72 | 0.40 | 0.81 | 0.76 | 0.62 | 0.42 | 0.58 | 0.26 | 0.65 | 0.09 | 0.59 | 0.32 | 0.53 | 0.27 | 0.71 | 0.41 | 0.53 | 0.30 |
| Ancestry | 0.52 | 0.21 | 0.47 | 0.30 | 0.43 | 0.28 | 0.58 | 0.38 | 0.68 | 0.39 | 0.72 | 0.59 | 0.53 | 0.38 | 0.82 | 0.57 | 0.61 | 0.41 |
| Education | 0.58 | 0.35 | 0.48 | 0.44 | 0.51 | 0.37 | 0.68 | 0.51 | 0.65 | 0.19 | 0.67 | 0.54 | 0.55 | 0.41 | 0.79 | 0.68 | 0.56 | 0.43 |
| Employment | 0.67 | 0.38 | 0.54 | 0.35 | 0.68 | 0.54 | 0.75 | 0.60 | 0.66 | 0.15 | 0.76 | 0.65 | 0.68 | 0.65 | 0.75 | 0.59 | 0.72 | 0.65 |
| Household | 0.69 | 0.43 | 0.73 | 0.68 | 0.60 | 0.54 | 0.64 | 0.43 | 0.67 | 0.26 | 0.71 | 0.58 | 0.66 | 0.60 | 0.79 | 0.59 | 0.68 | 0.64 |
| Housing | 0.71 | 0.66 | 0.58 | 0.59 | 0.67 | 0.63 | 0.74 | 0.71 | 0.68 | 0.36 | 0.74 | 0.65 | 0.70 | 0.66 | 0.71 | 0.72 | 0.73 | 0.66 |
| Income | 0.75 | 0.64 | 0.72 | 0.62 | 0.71 | 0.68 | 0.77 | 0.70 | 0.68 | 0.32 | 0.77 | 0.73 | 0.75 | 0.76 | 0.81 | 0.73 | 0.77 | 0.76 |
| Origin | 0.52 | 0.28 | 0.27 | 0.22 | 0.36 | 0.28 | 0.52 | 0.32 | 0.62 | 0.20 | 0.55 | 0.38 | 0.39 | 0.27 | 0.66 | 0.46 | 0.43 | 0.30 |
| Race | 0.57 | 0.40 | 0.45 | 0.37 | 0.42 | 0.27 | 0.54 | 0.34 | 0.67 | 0.14 | 0.76 | 0.65 | 0.57 | 0.48 | 0.83 | 0.64 | 0.64 | 0.54 |
| Residency | 0.51 | 0.03 | 0.37 | 0.19 | 0.35 | 0.01 | 0.49 | <0 | 0.62 | <0 | 0.46 | <0 | 0.33 | 0.04 | 0.59 | 0.19 | 0.40 | <0 |
| All | 0.84 | 0.81 | 0.86 | 0.88 | 0.80 | 0.79 | 0.83 | 0.82 | 0.85 | 0.76 | 0.88 | 0.86 | 0.84 | 0.83 | 0.90 | 0.87 | 0.85 | 0.84 |
| Diff | 0.24±0.11 | | 0.09±0.07 | | 0.13±0.09 | | 0.21±0.12 | | 0.45±0.10 | | 0.16±0.11 | | 0.12±0.09 | | 0.19±0.11 | | 0.14±0.11 | |

Note: Column I+G in each outcome uses both image and geospatial information, and G uses only geospatial embedding. Row Diff shows mean ± standard deviation of difference between I+G and G across categories, not including row All.

SUPPLEMENTAL TABLE V
TOP-3 PERFORMANCE OF GCGVT-A WITH IMAGE AND TWO GEOSPATIAL CATEGORY COMBINATIONS

| Outcome | Image with Categories | $R^2$ | Difference |
|---|---|---|---|
| Arthritis | AgeSex+Housing | 0.83 | 0.02 |
| | AgeSex+Income | 0.81 | 0.00 |
| | AgeSex+Employment | 0.80 | -0.01 |
| Cancer | AgeSex+Ancestry | 0.85 | -0.03 |
| | AgeSex+Education | 0.84 | -0.04 |
| | AgeSex+Race | 0.84 | -0.04 |
| CHD | AgeSex+Income | 0.79 | 0.00 |
| | AgeSex+Housing | 0.79 | 0.00 |
| | AgeSex+Employment | 0.77 | -0.02 |
| COPD | Employment+Income | 0.81 | 0.00 |
| | AgeSex+Income | 0.80 | -0.02 |
| | AgeSex+Housing | 0.80 | -0.02 |
| Depression | Ancestry+Income | 0.78 | 0.02 |
| | Income+Race | 0.78 | 0.02 |
| | Ancestry+Housing | 0.77 | 0.01 |
| Diabetes | Income+Race | 0.86 | 0.00 |
| | Housing+Race | 0.84 | -0.02 |
| | Ancestry+Housing | 0.84 | -0.02 |
| Kidney | AgeSex+Income | 0.79 | -0.04 |
| | AgeSex+Housing | 0.78 | -0.05 |
| | Ancestry+Income | 0.77 | -0.06 |
| Obesity | Income+Race | 0.88 | 0.01 |
| | Education+Race | 0.88 | 0.01 |
| | Ancestry+Education | 0.87 | 0.00 |
| Stroke | Income+Race | 0.82 | -0.03 |
| | AgeSex+Income | 0.80 | -0.04 |
| | Housing+Race | 0.79 | -0.05 |

Note: Column $R^2$ reports the performance of GCGVT-A using imagery and the two categories specified in each row. Difference is calculated as the row's $R^2$ minus the $R^2$ of GCGVT-A using all geospatial categories and no imagery.

SUPPLEMENTAL TABLE VI
EXAMPLES OF GENERATED SUMMARIES FROM GEOSPATIAL INFORMATION AND CORRESPONDING IMAGE

| Image | Generated Summary |
| --- | --- |
| 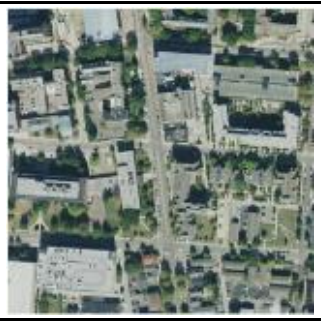 | This area predominantly comprises middle-aged to older adults, with a significant share of Hispanic and Asian populations, and a large number of suburban households, where a majority of part-time occupations are observed. |
| 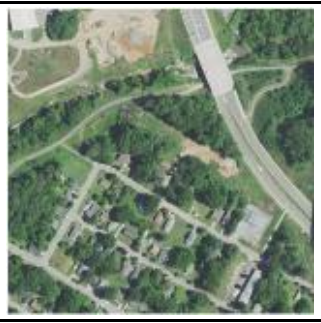 | This area exhibits significant disparities in socioeconomic status, with a majority of households belonging to the lower and very low income categories, and a substantial proportion of the population being African American or Hispanic. |
| 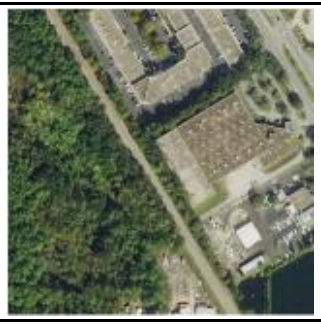 | This area exhibits a predominantly young population, particularly among householders, with a significant share of them being single, female, and elderly, and a notable absence of children. The majority of householders are native, and socioeconomically, they are primarily employed, with a large share working in the service sector. |
| 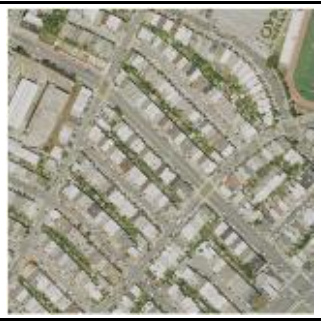 | This area predominantly consists of a significant number of households belonging to the Asian and White ethnicities, with a substantial minority of Hispanic households, particularly among the younger population. The majority of households are homeowners, and a considerable share of them are middle-aged or older. The employment status is primarily full-time, and the median income is above the national average. |
| 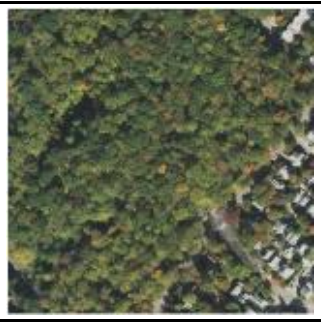 | This area predominantly consists of older households, with a significant minority of young households, and a large share of the population is native-born, particularly of Hispanic, Black, and White ethnicities, with a high proportion of married couples and a considerable number of homeowners. |
| 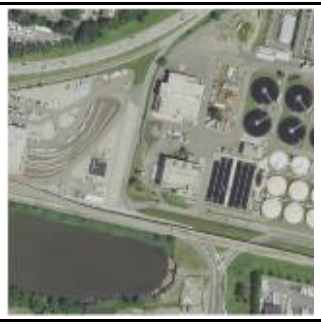 | This area predominantly consists of young households, with a significant share of renters, particularly among the lower-income population, and a high prevalence of commuting, with a majority of workers employed in the service sector. |
| 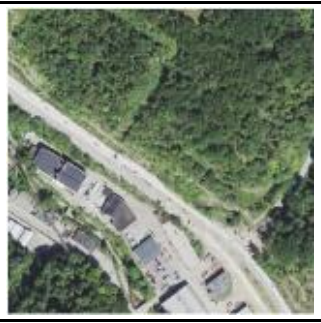 | This area predominantly comprises middle-aged to older households, with a significant minority of young adults, and a small share of children. The majority of households are homeowners, and a considerable portion of them are employed in the service sector. The median income is relatively low, and there is a high level of social assistance utilization. |
| 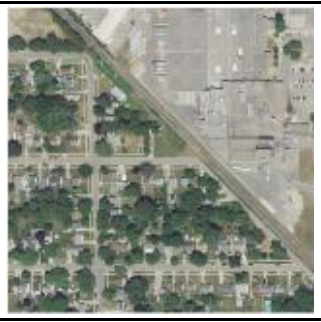 | This area predominantly comprises middle-aged homeowners, with a significant minority of retirees and a smaller share of young adults, particularly those in the labor force. The majority of homes are owned, and a substantial number are senior-occupied. |
| 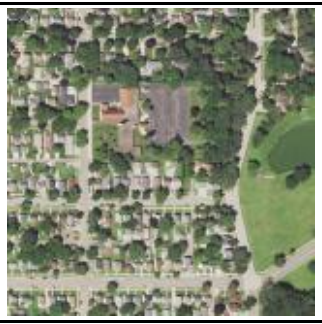 | This area is predominantly composed of middle-aged households, with a significant share of married or cohabiting couples, and a large proportion of females heading single-unit dwellings, particularly among older-aged units. The majority of households are homeowners, and a substantial number of them are Hispanic or African American. The most common occupation is in the service industry. |